\documentclass{article}

\usepackage{amsmath,amsfonts,bm}

\def\eqref#1{equation~\ref{#1}}
\def\1{\bm{1}}

\DeclareMathAlphabet{\mathsfit}{\encodingdefault}{\sfdefault}{m}{sl}
\SetMathAlphabet{\mathsfit}{bold}{\encodingdefault}{\sfdefault}{bx}{n}

\DeclareMathOperator*{\argmin}{arg\,min}

\usepackage{microtype}
\usepackage{graphicx}
\usepackage{subfigure}
\usepackage{booktabs}
\usepackage{url}
\usepackage{booktabs}
\usepackage{amsfonts}
\usepackage{nicefrac}
\usepackage{microtype}
\usepackage{wrapfig}
\usepackage[nointegrals]{wasysym}
\usepackage{lipsum}  
\usepackage{times}
\usepackage{tikz}
\usepackage{latexsym}
\usepackage{microtype}
\usepackage{graphicx}
\usepackage{placeins}
\usepackage{subcaption}
\usepackage{bbm}
\usepackage{multirow}
\usepackage{enumitem}
\usepackage{tikz}
\usepackage[export]{adjustbox}
\usepackage{amsmath}
\usepackage{amssymb}
\usepackage{mathtools}
\usepackage{amsthm}
\usepackage{nccmath}
\usepackage{algorithm}
\usepackage{caption}
\usepackage[algo2e, ruled]{algorithm2e}
\SetKwComment{Comment}{$\triangleright$\ }{}
\SetKwProg{Init}{Initialize}{}{}
\usepackage{textcomp,booktabs}
\usepackage{colortbl}
\definecolor{mygray}{gray}{.95}
\usepackage{multirow}
\usepackage{pifont}

\definecolor{definegray}{gray}{0.3}
\definecolor{definegreen}{HTML}{3FBC9D}

\usepackage{enumitem}
\usepackage{pifont}

\definecolor{CColor}{rgb}{0.01,0.31,0.59}
\definecolor{GGray}{rgb}{0.80,0.90,1}
\definecolor{Shady}{rgb}{0.9,0.9,0.9}
\definecolor{kaistblue}{RGB}{20,135,200}
\definecolor{kaistdarkblue}{RGB}{0,65,145}
\definecolor{urbanablue}{RGB}{19,41,75}
\definecolor{urbanaorange}{RGB}{232,74,39}
\definecolor{drp}{rgb}{0.53,0.15,0.34}
\usepackage[plainpages=false,pdfpagelabels,colorlinks=true,linkcolor=CColor,citecolor=CColor,urlcolor=CColor]{hyperref}

\newcommand{\methodname}{Surgery}

\usepackage{hyperref}

\usepackage[accepted]{icml2024}

\icmltitlerunning{Representation Surgery for Multi-Task Model Merging}

\begin{document}

\twocolumn[
\icmltitle{Representation Surgery for Multi-Task Model Merging}

\begin{icmlauthorlist}
\icmlauthor{Enneng Yang}{neu}
\icmlauthor{Li Shen}{sysu,jd}
\icmlauthor{Zhenyi Wang}{um}
\icmlauthor{Guibing Guo}{neu}
\icmlauthor{Xiaojun Chen}{szu}
\icmlauthor{Xingwei Wang}{neu}
\icmlauthor{Dacheng Tao}{ntu}
\end{icmlauthorlist}

\icmlaffiliation{neu}{Northeastern University, China.}
\icmlaffiliation{sysu}{Sun Yat-sen University, China.}
\icmlaffiliation{jd}{JD Explore Academy, China.}
\icmlaffiliation{um}{University of Maryland, USA.}
\icmlaffiliation{szu}{Shenzhen University, China.}
\icmlaffiliation{ntu}{Nanyang Technological University, Singapore}

\icmlcorrespondingauthor{Guibing Guo}{guogb@swc.neu.edu.cn}
%\icmlcorrespondingauthor{Enneng Yang}{ennengyang@stumail.neu.edu.cn}
\icmlcorrespondingauthor{Li Shen}{mathshenli@gmail.com}
%\icmlcorrespondingauthor{Zhenyi Wang}{wangzhenyineu@gmail.com}
% \icmlcorrespondingauthor{Guibing Guo}{guogb@swc.neu.edu.cn}
%\icmlcorrespondingauthor{Xiaojun Chen}{xjchen@szu.edu.cn}
%\icmlcorrespondingauthor{Xingwei Wang}{wangxw@swc.neu.edu.cn}
%\icmlcorrespondingauthor{Dacheng Tao}{dacheng.tao@gmail.com}

\icmlkeywords{Model Merging, Multi-task Learning}

\vskip 0.3in
]

\printAffiliationsAndNotice{} 

\begin{abstract}
Multi-task learning (MTL) compresses the information from multiple tasks into a unified backbone to improve computational efficiency and generalization. Recent work directly merges multiple independently trained models to perform MTL instead of collecting their raw data for joint training, greatly expanding the application scenarios of MTL. However, by visualizing the representation distribution of existing model merging schemes, we find that the merged model often suffers from the dilemma of \textit{representation bias}. That is, there is a significant discrepancy in the representation distribution between the merged and individual models, resulting in poor performance of merged MTL. In this paper, we propose a \textit{representation surgery} solution called ``\texttt{Surgery}" to reduce representation bias in the merged model. Specifically, Surgery is a lightweight task-specific module that takes the representation of the merged model as input and attempts to output the biases contained in the representation from the merged model. We then designed an unsupervised optimization objective that updates the Surgery module by minimizing the distance between the merged model's representation and the individual model's representation. Extensive experiments demonstrate significant MTL performance improvements when our Surgery module is applied to state-of-the-art (SOTA) model merging schemes. 
The code is available at 
% \href{https://github.com/EnnengYang/RepresentationSurgery}{RepresentationSurgery}.
\url{https://github.com/EnnengYang/RepresentationSurgery}.
\end{abstract}

\section{Introduction}
\label{sec:introduction}

Multi-task learning (MTL) utilizes a shared backbone to accommodate knowledge from multiple tasks simultaneously~\cite{SharedBottom_1997,mtlsurvey_tpami2021}. The MTL model is very attractive when considering model efficiency because it does not require saving a copy of the parameters for each task. Taking advantage of this, MTL have been widely used in computer vision~\cite{mtlasmooSenerK18_neurips2018,dwa_cvpr2019,Adashare_NeurIPS2020,CAGrad_NeurIPS2021,chen2022cerberus,adatask_AAAI2023}, natural language processing~\cite{mtl_nlp_icml2008,mtl_nlp_translation_acl2015,liu2016recurrent}, recommendation systems~\cite{mmoe_kdd2018,hadash2018rank,pan2019predicting,ple_recsys2020,wang2023multi,MMFI_TKDD_2024}, robotics~\cite{deisenroth2014multi,shridhar2023perceiver} and other fields~\cite{he2011graphbased,ishihara2021multi,mir_multiobjective,wu2023pi}.
However, MTL relies on the paradigm of ``collecting data first and then jointly training". 
This usually involves high data management costs and the risk of data privacy leakage. In addition, training the MTL model simultaneously also lacks flexibility because, when new tasks come, they need to be re-trained with all old tasks, which requires extra cost.

Recently, ``model fusion/merging" has emerged in the machine learning community~\cite{Modelsoups_ICML2022,FisherMerging_NeurIPS2022,RegMean_ICLR2023,TaskArithmetic_ICLR2023,TangentSpace_NeurIPS2023,TiesMerging_NeurIPS2023,pem_neurIPS2023,LoraHub_Arxiv2023,LinearizationLoRA_ICLR2024,AdaMerging_Arxiv2023}, which attempts to directly merge multiple independently trained or fine-tuned models to perform MTL~\cite{modelfusion_survey2023}, as shown in Fig.~\ref{fig:method}(a) to (b). In other words, model merging only requires trained model parameters for the respective tasks, that is, it no longer requires centralized management of MTL training data and joint training. In general, these methods greatly expand the application scenarios of MTL. Unfortunately, there is still a huge performance gap between the most advanced model merging based MTL~\cite{TaskArithmetic_ICLR2023,RegMean_ICLR2023,TiesMerging_NeurIPS2023,AdaMerging_Arxiv2023} and traditional MTL or individual models. This motivates us to further explore the existing problem in model merging and further solve it to close the above gap.

In this paper, we revisit several advanced model merging schemes~\cite{TaskArithmetic_ICLR2023,TiesMerging_NeurIPS2023,AdaMerging_Arxiv2023} from a \textbf{representation bias} perspective. Recall that the goal of model merging is to make the merged single model have the representational capabilities of multiple individual trained models. To examine the extent to which the merged model retains the representational capabilities of the original individual model, we conduct extensive experiments across eight tasks, three architectures, and four representative model merging methods. Specifically, we visualize (see Fig.~\ref{fig:dis_wo_surgery}) the distribution of representations extracted by the individual model (blue points) and the distribution of representations extracted by the merged model (red points) through t-SNE~\cite{tsne2008}. We observe: (i) There is a clear discrepancy between the two distributions, and this ``representation bias" exists \textit{across tasks, across architectures, and across model merging methods}. (ii) From Weight Averaging to Task Arithmetic~\cite{TaskArithmetic_ICLR2023} / Ties-Merging~\cite{TiesMerging_NeurIPS2023} to AdaMerging~\cite{AdaMerging_Arxiv2023}, the discrepancy between the two distributions is decreasing, which also corresponds to the performance improvement of the merged model. This indicates that the ``representation bias" problem is one of the biggest obstacles to model merging, and it also directly causes the poor performance of the merged MTL model.

To solve the ``representation bias" problem in model merging, we propose a novel \textbf{representation surgery} solution, abbreviated as ``\texttt{Surgery}", to filter out representation bias from other tasks after model merging. Specifically, \texttt{Surgery} is a task-specific lightweight module that takes the representation extracted by the merged model as input and attempts to filter the bias contained in it, so that the filtered representation output is as close as possible to the feature representation of an individual model. Since raw training data is unavailable, we design an unsupervised surrogate objective to update parameters in the \texttt{Surgery} module, which minimizes the distance between the representation after surgery and the representation of the individual model. Our \texttt{Surgery} is completely orthogonal to existing model merging solutions, and it can be incorporated into any model merging method to solve its representation bias problem. We conduct extensive experiments on eight tasks and three architectures, and the results show that when our \texttt{Surgery} module is applied to several advanced model merging schemes, the performance of the merged MTL model can be significantly improved. 

The main contributions of this paper are three-fold:
\begin{itemize}[noitemsep,topsep=0pt,parsep=0pt,partopsep=0pt,leftmargin=*]
    \item We revisit SOTA model merging methods and identify for the first time the ``representation bias" problem (which exists across tasks, architectures, and merging methods) as a major cause of poor MTL performance. 
    \item We propose a novel ``representation surgery" approach, called \texttt{Surgery}, to solve ``representation bias" in the merged model. Moreover, the \texttt{Surgery} module is suitable for any existing model merging algorithm.
    \item We conduct extensive experiments to verify that our \texttt{Surgery} scheme can effectively alleviate representation bias and significantly improve MTL performance.
\end{itemize}

\section{Related Work}
\label{sec:relatedwork}

\subsection{Model Merging for Multi-Task Learning}
\label{subsec:relatedwork_modelmerging}

Model merging is to merge multiple individual models into one~\cite{modelfusion_survey2023}. It has two main application scenarios: First, merge models trained on the \textit{same task} to improve the accuracy or generalization of the final model~\cite{SWA_ICLR2020,SWAD_NeurIPS2021,Modelsoups_ICML2022,lu2022improving_EMNLP2022,TWA_ICLR2023}. 
Second, merge multiple models trained on \textit{different tasks} to perform MTL~\cite{TaskArithmetic_ICLR2023,TiesMerging_NeurIPS2023,rewardedsoups_neurips2023,pem_neurIPS2023,ZipIt_Arxiv2023,LoraHub_Arxiv2023,AdaMerging_Arxiv2023}, which is the focus of this paper. We further divide model merging into two stages: before and during merging. 

(i) The main concern \textit{before merging} is how to provide more favorable preconditions for model merging, such as linearization or orthogonalization. Specifically, \cite{TangentSpace_NeurIPS2023} independently fine-tunes each task in the Tangent space~\cite{NTK_NeurIPS2018} of the pre-trained model and demonstrates that this helps decouple the weight space from the input space, leading to better model merging. Similarly, Linearization-LoRA~\cite{LinearizationLoRA_ICLR2024} linearly fine-tunes some LoRA modules~\cite{lora_iclr2022} in Tangent space.
In addition, Task Arithmetic~\cite{TaskArithmetic_ICLR2023} pointed out that the orthogonality between task vectors is one of the conditions for successful model merging.

(ii) The main focus \textit{during merging} is how to mitigate interference and conflicts between models~\cite{TaskArithmetic_ICLR2023,TiesMerging_NeurIPS2023,MergeLM_Arxiv2023,ReBasin_CVPR2023,RegMean_ICLR2023,rewardedsoups_neurips2023,pem_neurIPS2023,AdaMerging_Arxiv2023}. For example, Ties-Merging~\cite{TiesMerging_NeurIPS2023} eliminates the problem of parameter sign conflicts during model merging. DARE~\cite{MergeLM_Arxiv2023} removes a large number of useless neuron updates and then scales the neurons for merging. ReBasin~\cite{ReBasin_ICLR2023,ReBasin_CVPR2023} rearranges and aligns the neurons of multiple models to establish connectivity paths between multiple models. Fisher-Merging~\cite{FisherMerging_NeurIPS2022} performs weighted merging utilizing the importance of each parameter through the Fisher information matrix~\cite{fisher1922mathematical}. RegMean~\cite{RegMean_ICLR2023} reweights and linearly combines rows in weight matrices based on statistics from training data. Concrete~\cite{tang2023concrete} finds a shared subspace between multiple tasks for model merging.
AdaMerging~\cite{AdaMerging_Arxiv2023} leverages unlabeled test data to automatically learn a set of task- or layer-level model merging coefficients.

While existing methods predominantly concentrate on merging in weight space, they often neglect a crucial concern stemming from weight merging--the \textit{representation bias}. A substantial disparity emerges in the \textit{representation space} between the merged model and individually-trained models. In contrast, our surgery method addresses this gap, aiming to minimize the representation discrepancy. Moreover, our approach operates in the \textit{representation space}, offering a complementary and orthogonal perspective to traditional weight-space merging methods. Consequently, our method can be seamlessly integrated with them.

\subsection{Traditional Multi-Task Learning}
\label{subsec:relatedwork_mtl}

MTL usually faces negative transfer~\cite{SharedBottom_1997,mtlsurvey_tpami2021,negative_transfer_survey_2022}.
Existing work mainly solves the negative transfer problem from two directions: architecture and optimization. Specifically,
(i) The classic SharedBottom~\cite{SharedBottom_1997} backbone performs poorly when tasks are not highly correlated. Advanced \textit{{architectures}} primarily alleviate the phenomenon of negative transfer through modularization~\cite{mmoe_kdd2018,SNR_AAAI2019,ple_recsys2020}, sparsification~\cite{Adashare_NeurIPS2020,dwa_cvpr2019}, and soft sharing~\cite{CrossStitch_CVPR2016,mtlnas_CVPR2020} of the backbone. 
(ii) Other work alleviates task interference from an \textit{{optimization}} perspective. For example, \cite{uncertaintyweighting_cvpr2018,mtlasmooSenerK18_neurips2018,dwa_cvpr2019,liu2022auto_lambda,revisitingScalarization_NeurIPS2023} try to optimize the weight for each task loss.  \cite{graddrop_neurips2020,PCGrad_NeurIPS2020,CAGrad_NeurIPS2021,wang2021gradientVaccine,javaloy2022rotograd,nashmtl_ICML2022} resolve multi-task gradient direction or sign conflicts. \cite{gradnorm_icml2018,imtl_iclr2021,MetaBalanceWWW2022,adatask_AAAI2023} try to eliminate the dominance of the learning rate or gradient.
Distinct from these existing traditional MTL approaches that concentrate on loss weight or gradient space to tackle the negative transfer issue, our method improves MTL performance by addressing the representation bias problem within the representation space in model merging based MTL, providing a novel and orthogonal perspective.

\section{Representation Bias in Model Merging}
\label{sec:method}

We first give the notation and definition in Sec.~\ref{subsec:preliminaries}. Next, we revisit existing model merging schemes in Sec.~\ref{subsec:rethinking} and point out their existing ``representation bias'' issues.

\subsection{Preliminaries}
\label{subsec:preliminaries}

\textbf{Notations}: Denote the neural network model $\small f: \mathcal{X} \times \Theta \mapsto \mathcal{Y}$, the parameter is $\small \mathbf{\theta} \in {\Theta} \in \mathbb{R}^n$, the input is $\small \mathbf{x}_i \in \mathcal{X}  \in \mathbb{R}^d$, and the output is $\small \mathbf{y}_i \in \mathcal{Y}  \in \mathbb{R}^c$. Among them, $n$ represents the number of parameters, $d$ represents the dimension of the input data, and $c$ represents the number of output classes. Considering $T$ independently fine-tuned models $f_{\theta_t}$ (where $\small t \in \{1, 2, \ldots, T\}$), $\small f_{\theta_t}$ is well trained on their respective training data $\small \mathcal{D}^t_{tr}(\mathcal{X}, \mathcal{Y})$. 
Without loss of generality, it is usually assumed that $\small \{f_{\theta_t}\}_{t=1}^T$ are all fine-tuned on a popular backbone $\small f_{\theta_0}$ (i.e., the pre-trained weight is $\theta_0$), such as ResNet~\cite{resnet}, ViT~\cite{Vit_ICLR2021} or BERT~\cite{BERT_NAACL2019}.

\textbf{Problem Setup}: Model merging is to merge the weights $\small \{\theta_t\}_{t=1}^T$ to obtain a final weight $\small {\theta}_{mtl}^m$, so that $\small f_{{\theta}_{mtl}^m}$ can simultaneously perform the tasks $\small \{1,2,\ldots,T\}$, where $m$ is a model merging method. Additionally, we are not allowed to access raw training data $\small \mathcal{D}^t_{tr}$ when merging models. Formally, we expect the loss of the merged model $\small f_{{\theta}_{mtl}^m}$ to be as small as possible on the test dataset $\small \{\mathcal{D}^t_{te}\}^T_{t=1}$  of all tasks, i.e., $\small \min \frac{1}{T}\sum_{t=1}^T\sum_{i=1}^{|\mathcal{D}^t_{te}|} \frac{1}{|\mathcal{D}^t_{te}|} \ell (f_{\theta_{mtl}^m}(\mathbf{x}_i), \mathbf{y}_i)$, where $\small \ell(\cdot)$ is the loss function, such as the cross-entropy loss.

\begin{figure*}[t]
    \centering 
    \subfigure{
        \begin{minipage}[t]{0.48\linewidth}
         \includegraphics[width=.49\textwidth]{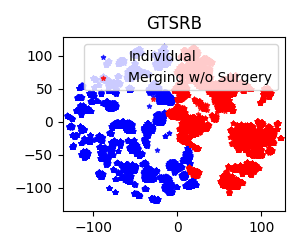
        }
        \includegraphics[width=.49\textwidth]{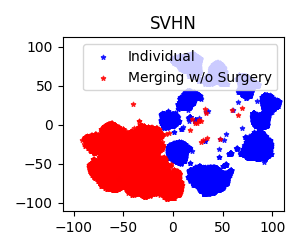
        }
        \vspace{-10pt}
        \begin{center}
            \text{\small (a) Weight Averaging on ViT-B/32}
         \end{center}
        \end{minipage}
    }
    \vspace{-10pt}
    \subfigure{
        \begin{minipage}[t]{0.48\linewidth}
         \includegraphics[width=.49\textwidth]{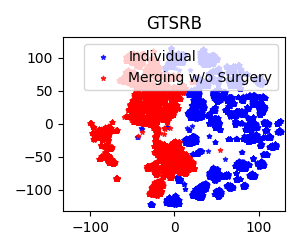}
        \includegraphics[width=.49\textwidth]{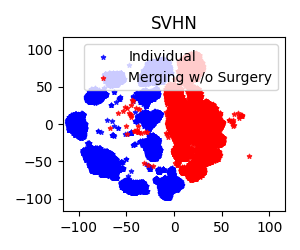}
        \vspace{-10pt}
        \begin{center}
            \text{\small (b) Task Arithmetic on ViT-B/32}
         \end{center}
        \end{minipage}
    }
    \vspace{-10pt}
    \subfigure{
        \begin{minipage}[t]{0.48\linewidth}
         \includegraphics[width=.49\textwidth]{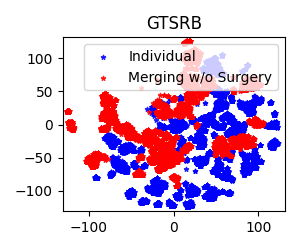}
     \includegraphics[width=.49\textwidth]{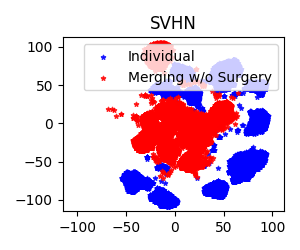}
        \vspace{-10pt}
        \begin{center}
            \text{\small (c) AdaMerging on ViT-B/32}
         \end{center}
        \end{minipage}
    }
    % \vspace{-10pt}
    \subfigure{
        \begin{minipage}[t]{0.48\linewidth}
         \includegraphics[width=.49\textwidth]{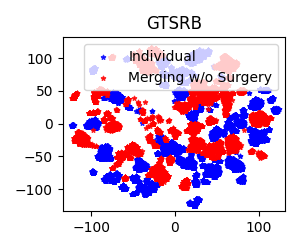}
        \includegraphics[width=.49\textwidth]{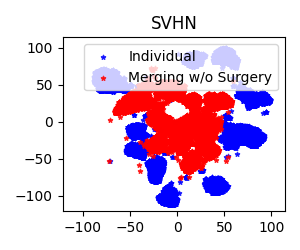}
        \vspace{-10pt}
        \begin{center}
            \text{\small (d) AdaMerging on ViT-B/16}
         \end{center}
        \end{minipage}
    }
    \caption{Visualization of the distribution of representations extracted by the merged model (\textcolor{red}{red}) for the existing model merging schemes and representations extracted by the individual model (\textcolor{blue}{blue}). We observe that there is a clear distribution discrepancy between the two.}  
\label{fig:dis_wo_surgery} 
\vspace{-12pt}
\end{figure*}

\begin{figure*}[t]
    \centering 
    \includegraphics[width=0.48\textwidth]{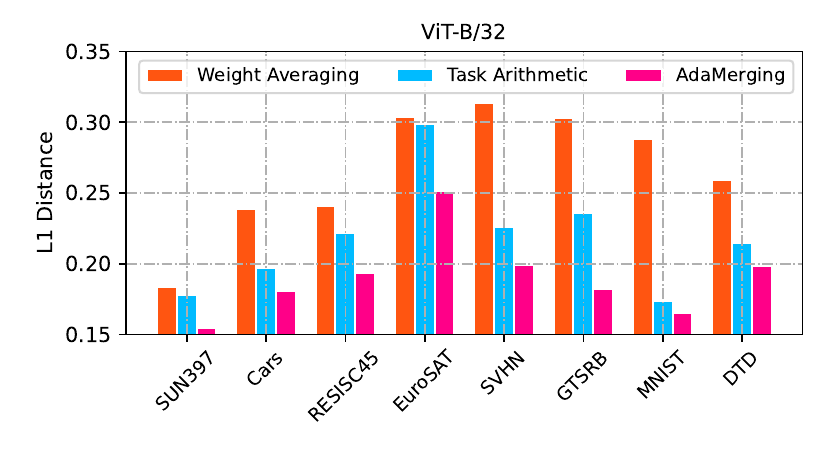}
    \includegraphics[width=0.48\textwidth]{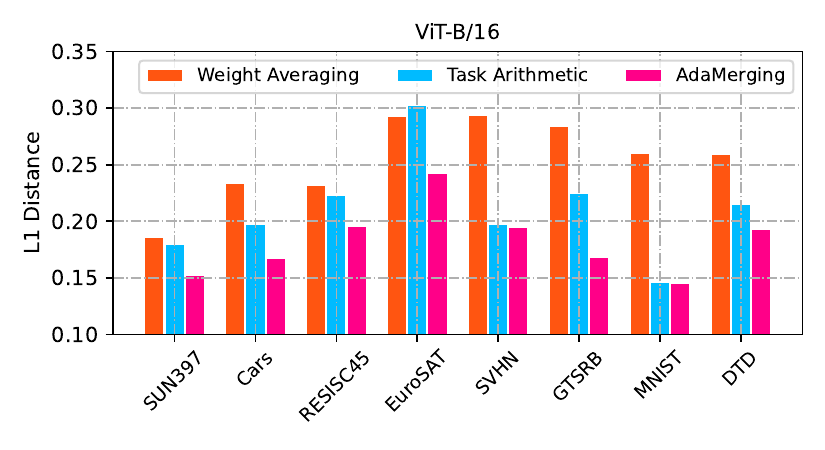}
    \vspace{-18pt}
    \caption{The $L_1$ distance (or ``representation bias'' in Eq.~\ref{eq:l1distance}) between representations extracted by the merged model using various model merging methods (i.e., Weight Averaging, Task Arithmetic and AdaMerging) and representations extracted by individual models. 
    }  
\label{fig:l1_distance_compare} 
\vspace{-10pt}
\end{figure*}

\textbf{Representative Model Merging Solutions}: 
\ding{192} The simplest merging scheme is \textit{\textbf{Weight Averaging}}, which directly averages the parameters $\small \{\theta_t\}_{t=1}^T$ of multiple models: $\small \theta_{mtl}^m = \frac{1}{T} \sum_{t=1}^T \theta_t$. However, the performance of simple weighted averaging is often unsatisfactory. 
\ding{193} Recently, \textit{\textbf{Task Arithmetic}}~\cite{TaskArithmetic_ICLR2023} record the parameter difference ($\small \tau_t = \theta_t - \theta_0$) obtained by subtracting the pre-trained model $\small f_{\theta_0}$ from the fine-tuned model $\small f_{\theta_t}$ as the \textit{task vector}. Then, it merges the task vectors $\small \{\tau_t\}_{t=1}^T$ of multiple tasks and combines them into the pre-trained weight $\small \theta_0$: $\small \theta_{mtl}^m = \theta_0 + \lambda \sum_{t=1}^T \tau_t$. By simply adjusting the hyperparameter $\lambda$, Task Arithmetic can achieve better model merging performance than Weight Averaging. 
\ding{194} On the basis of task vector, \textit{\textbf{Ties-Merging}}~\cite{TiesMerging_NeurIPS2023} further proposes three operations: TRIM, ELECT SIGN and MERGE to eliminate the symbol conflict problem in task vectors. We combine these three operations and call them $\small \phi(\cdot)$. Finally, the Ties-Merging merge is expressed as: $\small \theta_{mtl}^m = \theta_0 + \lambda \sum_{t=1}^T \phi(\tau_t)$. 
\ding{195} Furthermore, \textit{\textbf{AdaMerging}}~\cite{AdaMerging_Arxiv2023} adaptively learns a set of task-level or layer-level merging coefficients for Task Arithmetic or Ties-Merging, which significantly improves the MTL performance of model merging. Task-wise and layer-wise AdaMerging are expressed as: $\small \theta_{mtl}^m = \theta_0 + \lambda_t \sum_{t=1}^T \tau_t$ and $\small \theta_{mtl}^m = \{\theta^l_0 + \lambda_t^l \sum_{t=1}^T \tau^l_t\}_{l=1}^L$ (where $L$ is the number of layers in model $\small f_{\theta_{mtl}^m}$) respectively. For other mature model merging solutions, please refer to Sec.~\ref{subsec:relatedwork_modelmerging}.

\begin{figure*}[t]
    \centering 
    \includegraphics[width=.95\textwidth]{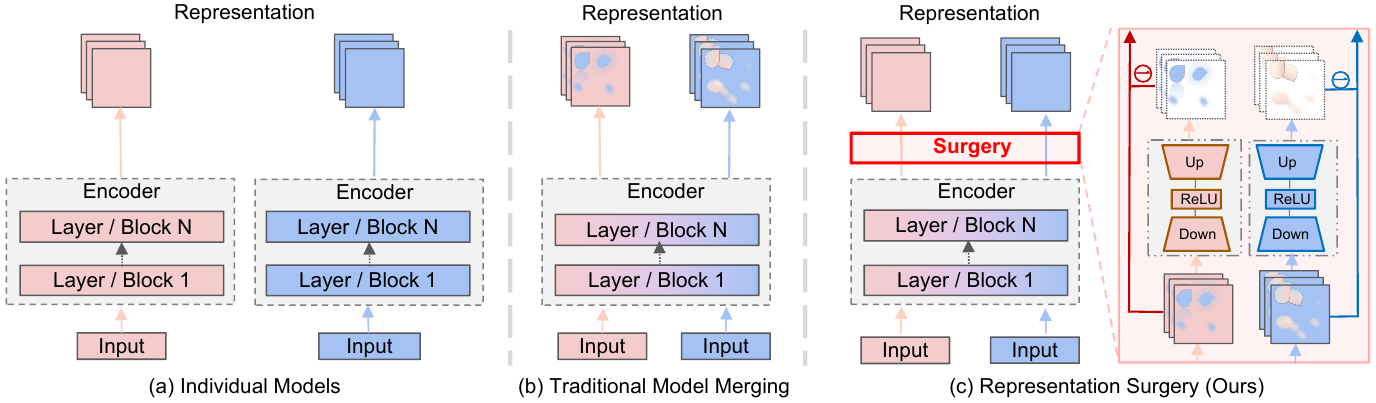}
    \vspace{-5pt}
    \caption{Representation Surgery for Multi-Task Model Merging. (a) Multiple individual trained models. (b) Traditional model merging schemes (e.g., Task Arithmetic~\cite{TaskArithmetic_ICLR2023}, Ties-Merging~\cite{TiesMerging_NeurIPS2023}, AdaMerging~\cite{AdaMerging_Arxiv2023}, etc.) merge multiple individual models into one. However, they usually suffer from the ``representation bias'' problem. (c) ``Representation surgery'' solution is proposed in this paper. It is a task-specific lightweight module used to solve the representation bias problem.}  
\label{fig:method} 
\vspace{-10pt}
\end{figure*}

\subsection{Revisiting Representation Bias}
\label{subsec:rethinking}

This section revisits existing model merging schemes and finds that they suffer from a ``representation bias'' problem, that is, the feature representations extracted by the merged model differ from those of the individual models.

\subsubsection{Settings}
\label{subsubsec:setting}

Without loss of generality, we perform analysis on the following representative model merging methods, tasks/datasets, and architectures. 
(i) \textit{\textbf{Methods}}: We analyze the four model merging methods mentioned in Sec.~\ref{subsec:preliminaries}: Weight Averaging, Task Arithmetic, Ties-Merging and AdaMerging.
(ii) \textit{\textbf{Tasks}}: We follow Task Arithmetic~\cite{TaskArithmetic_ICLR2023} and use the following eight datasets as eight tasks for model merging: SUN397, Cars, RESISC45, EuroSAT, SVHN, GTSRB, MNIST, DTD. We provide a detailed dataset description in Appendix~\ref{sec:experimentalsetting_appendix}.
(iii) \textit{\textbf{Architectures}}: We merge eight models into one model and experiment with three ViT architectures~\cite{Vit_ICLR2021} with different parameter scales: ViT-B/32, ViT-B/16, and ViT-L/14.

To explore the reasons for the performance gap between the merged model $\small f_{\theta_{mtl}^m}$ and individual models $\small \{f_{\theta_t}\}_{t=1}^T$, we tried to visualize the distribution of the feature representations they extracted. Specifically, we first input all the test samples $\small D^t_{te}(\mathcal{X}, \mathcal{Y})$ of each task $t$ into the models $\small f_{\theta_{mtl}^m}$ and $\small f_{\theta_t}$ respectively, and record their extracted feature representations (i.e., the \textit{\textbf{final layer}} before task-specific head/heads of the individual/merged models) as $\small \mathcal{Z}^{mtl}_{t} \in \mathbb{R}^{N \times k}$ and $\small \mathcal{Z}^{ind}_{t} \in \mathbb{R}^{N \times k}$ respectively. Among them, $\small N=|D^t_{te}|$ represents the amount of data, and $k$ represents the dimension of the feature (e.g., ViT-B/32 and ViT-B/16 are 512, and ViT-L/14 is 768).
Next, we can map the high-dimensional data $\small \mathcal{Z}^{mtl}_{t}$ and $\small \mathcal{Z}^{ind}_{t}$ to a lower-dimensional space (2-dimensional) through t-SNE tool and visualize it.

\subsubsection{``Representation Bias'' Problem}
\label{subsubsec:representation_bias}

We find that ``representation bias'' exists \textit{across tasks, across merging methods, and across architectures}~\footnote{\footnotesize Due to space limitations, we only show two tasks (GTRSB and SVHN), three merging methods (Weight Averaging, Task Arithmetic and AdaMerging), and two architectures (ViT-B/32 and ViT-B/16) in the main text. The comprehensive results are presented in Appendix~\ref{sec:performance_appendix} (A summary is given in Tab.~\ref{tab:overall_distribution}).}. 
Specifically, as shown in  Fig.~\ref{fig:dis_wo_surgery}, we have the following observations: 
(i) ``Representation bias'' is present across various tasks. As shown in Fig.~\ref{fig:dis_wo_surgery}(a), on the two tasks of GTSRB and SVHN, the feature representation distributions of the merged model (red points) and the individual model (blue points) are quite different. As shown in Appendix~\ref{sec:performance_appendix}, this representation bias problem also exists in the other six tasks.
(ii) ``Representation bias'' persists across different merging methods. Comparing Fig.~\ref{fig:dis_wo_surgery}(a), (b), (c), we find that the phenomenon of inconsistent representation distribution between the merged model and the individual model exists in Weight Averaging, Task Arithmetic and AdaMerging.
(iii) ``Representation bias'' exists across diverse architectures. Comparing Fig.~\ref{fig:dis_wo_surgery}(c) and (d), we find that the representation bias problem holds true in both ViT-B/32 and ViT-B/16, and the results in Appendix~\ref{sec:performance_appendix} show that it also holds true in ViT-L/14.

This phenomenon encourages us to think further. \textit{Is representation bias the key factor limiting the performance improvement of model merging}? We give a ``yes" answer.
\textbf{First}, a large amount of previous experimental experience~\cite{TaskArithmetic_ICLR2023,TiesMerging_NeurIPS2023,AdaMerging_Arxiv2023} shows that in terms of model merging performance, AdaMerging $\small >$ Task Arithmetic $\small >$ Weight Averaging, the results are shown in Tab.~\ref{tab:performance_vitbase32}, Tab.~\ref{tab:performance_vitlarge14} and Tab.~\ref{tab:performance_vitbase16_appendix} (in Appendix~\ref{sec:performance_appendix}). For example, in Tab.~\ref{tab:performance_vitbase32}, the average accuracy of AdaMerging, Task Arithmetic, and Weight Averaging on eight tasks are 80.1\%, 69.1\% and 65.8\% respectively.
\textbf{Second}, by directly observing the discrepancy between the red and blue distributions in Fig.~\ref{fig:dis_wo_surgery}(a)(b)(c), we can see the phenomenon of AdaMerging $\small <$ Task Arithmetic $\small <$ Weight Averaging. Furthermore, to quantitatively discuss the distance (or \textit{representation bias}) between the representation of each merged model (e.g., by AdaMerging, Task Arithmetic, Weight Averaging) and that of individual models, we calculated the $L_1$ distance between feature representations $\mathcal{Z}^{mtl}_{t}$ and $\mathcal{Z}^{ind}_{t}$ in Sec.~\ref{subsubsec:setting} on the test dataset $D^t_{te}(\mathcal{X}, \mathcal{Y})$ of each task $t$, i.e., 
\begin{equation}
\small
    \text{representation bias: } d_t=\frac{1}{k} \frac{1}{|D^t_{te}|}  \|\mathcal{Z}^{mtl}_{t} - \mathcal{Z}^{ind}_{t} \|_1.
\label{eq:l1distance}
\end{equation} 

As shown in Fig.~\ref{fig:l1_distance_compare}, we clearly observe that AdaMerging has a smaller representation bias between the merged model and the individual models compared to the other two model merging methods, i.e., Task Arithmetic and Weight Averaging. Meanwhile, Task Arithmetic also has a smaller representation bias than Weight Averaging.

The above analysis shows that representation bias poses significant challenges when completing MTL based on model merging. The performance of the merged model improves when the representation bias decreases. This motivates us to seek a solution to alleviate ``representation bias" problem in model merging based MTL methods.

\section{Representation Surgery for Model Merging}
\label{subsec:surgery}

In this section, we propose a simple yet effective method-agnostic representation surgery scheme to alleviate the representation bias problem pointed out in the above section.

\begin{figure*}[h]
    \centering 
    \subfigure{
        \begin{minipage}[t]{0.48\linewidth}
         \includegraphics[width=.49\textwidth]{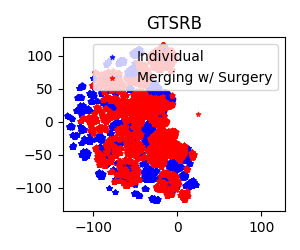
        }
        \includegraphics[width=.49\textwidth]{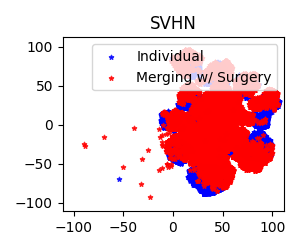
        }
        \vspace{-10pt}
        \begin{center}
            \text{\small (a) Weight Averaging on ViT-B/32}
         \end{center}
        \end{minipage}
    }
    \vspace{-10pt}
    \subfigure{
        \begin{minipage}[t]{0.48\linewidth}
         \includegraphics[width=.49\textwidth]{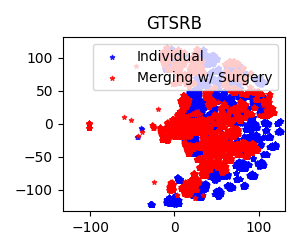}
        \includegraphics[width=.49\textwidth]{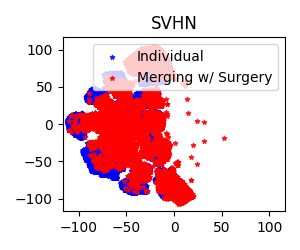}
        \vspace{-10pt}
        \begin{center}
            \text{\small (b) Task Arithmetic on ViT-B/32}
         \end{center}
        \end{minipage}
    }
    \vspace{-10pt}
    \subfigure{
        \begin{minipage}[t]{0.48\linewidth}
         \includegraphics[width=.49\textwidth]{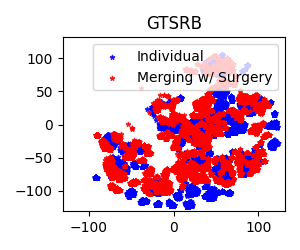}
     \includegraphics[width=.49\textwidth]{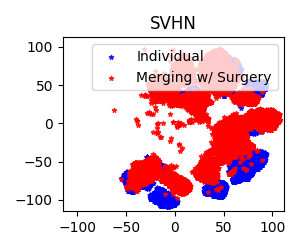}
        \vspace{-10pt}
        \begin{center}
            \text{\small (c) AdaMerging on ViT-B/32}
         \end{center}
        \end{minipage}
    }
    \subfigure{
        \begin{minipage}[t]{0.48\linewidth}
         \includegraphics[width=.49\textwidth]{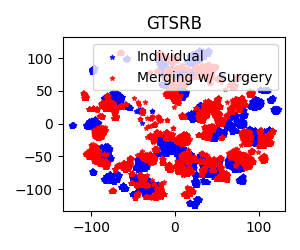}
        \includegraphics[width=.49\textwidth]{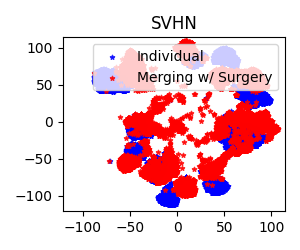}
        \vspace{-10pt}
        \begin{center}
            \text{\small (d) AdaMerging on ViT-B/16}
         \end{center}
        \end{minipage}
    }
    \caption{Visualization of the distribution of features extracted by the merged model {\textbf{after performing the representation surgery}} (\textcolor{red}{red}) and features extracted by the individual model (\textcolor{blue}{blue}). We observe that the distributions of the two are relatively close.}  
\label{fig:dis_w_surgery} 
\vspace{-10pt}
\end{figure*}

\begin{figure*}[t]
    \centering 
    \includegraphics[width=.48\textwidth]{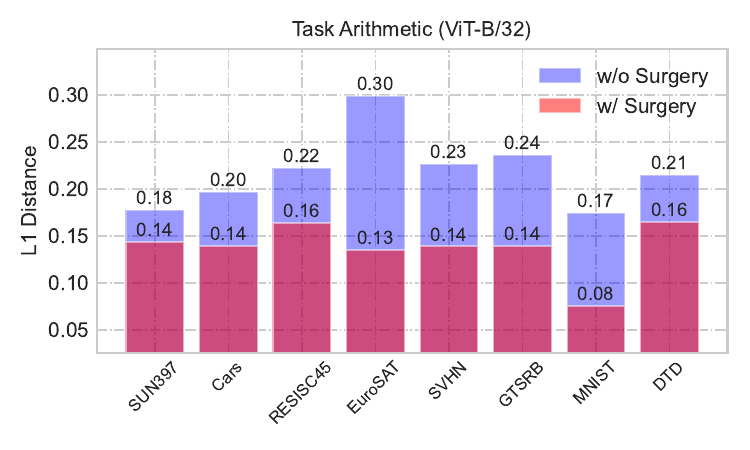}
    \includegraphics[width=.48\textwidth]{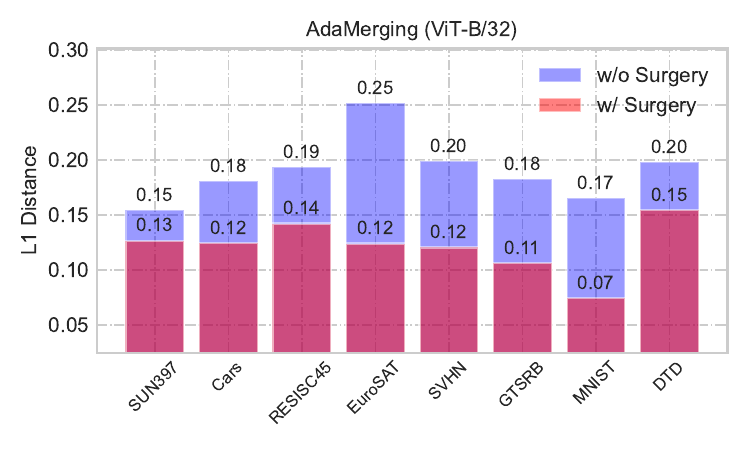}
    \vspace{-20pt}
    \caption{Visualization of the $L_1$ distance (or ``representation bias'' in Eq.~\ref{eq:l1distance}) of the representation of the merged model \textbf{with} (\textcolor{red}{red}) and \textbf{without} (\textcolor{blue}{blue}) representation surgery versus the individual model.
    }  
\label{fig:l1_distance_vitb32} 
\vspace{-10pt}
\end{figure*}

\subsection{Optimization Objective}
\label{subsubsec:Objective}

We directly take the representation bias (i.e., Eq.~\ref{eq:l1distance}) as the optimization goal of representation surgery, that is, minimizing the distance between the representation extracted by the merged model and the representation extracted by the individual model. In other words, as shown in Fig.~\ref{fig:method}(c), our representation surgery is to filter out the representation bias $\small \Phi_t(\mathcal{Z}^{mtl}_{t})$ of the representation $\small \mathcal{Z}^{mtl}_{t}$ extracted by the merged model $\small f_{\theta_{mtl}^m}$, thereby making it (i.e., $\small \mathcal{Z}^{mtl}_{t} - \Phi_t(\mathcal{Z}^{mtl}_{t})$) closer to the representation $\small \mathcal{Z}^{ind}_{t}$ of the individual model $\small f_{\theta_t}$. Formally, the optimization problem of representation surgery is as follows:
\begin{equation}
\small
\begin{split}
     \argmin_{\{\theta_{\Phi_1},\theta_{\Phi_2},\ldots,\theta_{\Phi_T}\}}  & \sum_{t=1}^T \frac{1}{|D^t_{te}|} \| \mathcal{\hat{Z}}^{mtl}_{t} - \mathcal{Z}^{ind}_{t} \|_1 \\
     \text{s.t. } &\mathcal{\hat{Z}}^{mtl}_{t} = \mathcal{Z}^{mtl}_{t} - \Phi_t(\mathcal{Z}^{mtl}_{t}), 
\end{split}
\label{eq:objective}
\end{equation}
where $\small \Phi_t(\cdot)$ is a task-private lightweight module, which can be an arbitrary implementation (such as multiple fully connected layers, etc.). 
Without loss of generality, in this paper, we implement it as an Adapter~\cite{adapter_icml2019}-like structure, that is, 
\begin{equation}
\small
    \Phi_t(\mathcal{Z}^{mtl}_{t})=\mathbf{W}_{up} \cdot \text{ReLU}(\mathbf{W}_{down} \cdot {\mathcal{Z}^{mtl}_{t}}^{\top})
\label{eq:surgery}
\end{equation}
where $\small \mathbf{W}_{up} \in \mathbb{R}^{k \times r}$, $\small \mathbf{W}_{down} \in \mathbb{R}^{r \times k}$ represent two learnable matrices, i.e., $\small \theta_{\Phi_t}=\{\mathbf{W}_{up}, \mathbf{W}_{down}\}$, ReLU($\cdot$) is a nonlinear activation function, $k$ is the dimension of representation as mentioned in Sec.~\ref{subsec:preliminaries}, and $r$ is a hyperparameter, also called rank, which we set to 16 by default.

Overall, our \texttt{Surgery} scheme {does not rely on any labeled training data}, but utilizes unlabeled test data  $\small \{\mathcal{D}^t_{te}(\mathcal{X}, \mathcal{Y})\}_{t=1}^T$ and individual models $\small \{f_{\theta_t}\}_{t=1}^T$ as a self-supervised signal to train the \texttt{Surgery} module's parameters $\small \{\theta_{\Phi_1},\theta_{\Phi_2},\ldots,\theta_{\Phi_T}\}$. In addition, as mentioned in Appendix~\ref{sebsec:analysis}, the number of parameters increased by our surgery module is very minor ($\small 0.01\%$) compared to the number of parameters that need to be merged into the model.

\begin{table*}[t]
\vspace{-5pt}
\centering
\caption{Multi-task performance when merging ViT-B/32 models on eight tasks.}
\label{tab:performance_vitbase32} 
\resizebox{\linewidth}{!}{  
\begin{tabular}{l|cccccccc|cc}
\toprule
{Method}  &  {SUN397}  &  {Cars}  &  {RESISC45}  &  {EuroSAT}  &  {SVHN}  &  {GTSRB}  &  {MNIST}  &  {DTD}  & \textbf{Avg.}  \\
\midrule
{Pretrained}  &  {62.3}  &  {59.7}  &  {60.7}  &  {45.5}  &  {31.4}  &  {32.6}  &  {48.5}  &  {43.8} & {48.0}  \\
{Individual}  &  75.3  &  77.7  &  96.1  &  99.7  &  97.5  &  98.7  &  99.7  &  79.4 & 90.5   \\
{Traditional MTL}    &  73.9  &  74.4  &  93.9  & 98.2    &  95.8  &  98.9   &  99.5   & 77.9 & 88.9  \\
\midrule
{Weight Averaging} & 65.3  &  63.4  &  71.4  &  71.7  &  64.2  &  52.8  &  87.5  &  50.1  & 65.8 \\
{Fisher Merging}~\citep{FisherMerging_NeurIPS2022}   &  68.6  &  69.2  &  70.7  &  66.4  &  72.9  &  51.1  &  87.9  &  59.9 & 68.3  \\
{RegMean}~\citep{RegMean_ICLR2023}   &  65.3  &  63.5  &  75.6  &  78.6  &  78.1  &  67.4  &  93.7  &  52.0 & 71.8 \\
\midrule
Task Arithmetic~\citep{TaskArithmetic_ICLR2023}   &55.2 &54.9 &66.7 &78.9 &80.2 & 69.7 &97.3 &50.4 & 69.1 \\
{Ties-Merging}~\citep{TiesMerging_NeurIPS2023}   &  65.0  & 64.4  & 74.8  &  77.4  &  81.2  & 69.3 &  96.5  &  54.5 & 72.9  \\
Concrete TA~\cite{tang2023concrete} & 62.5 &  61.1 & 76.0 & 95.7 &  91.0 &81.9 & 98.5& 51.9 & 77.3 \\
Concrete AM~\cite{tang2023concrete} &  67.8 &70.0 & 87.5 &  96.0 & 91.6 & 96.7 & 98.7 & 63.8 &84.0\\
{TW AdaMerging}~\cite{AdaMerging_Arxiv2023} &58.0 &53.2 &68.8 &85.7 &81.1 &84.4 &92.4  &44.8 &71.1 \\
{AdaMerging}~\cite{AdaMerging_Arxiv2023}   &64.5 &68.1 &79.2 &93.8 &87.0 &91.9 &97.5  &59.1 &80.1 \\
\midrule
\rowcolor{mygray}
\textbf{Weight Averaging w/ \methodname} (Ours) & 67.6 &64.6 &85.8 &96.8 &76.9 &82.9 &97.8 &67.3 &80.0\\
\rowcolor{mygray}
\textbf{Task Arithmetic w/ \methodname} (Ours)  &63.8 &59.9 &83.3 &97.9 &87.0 &87.0 &98.6 &69.4 &80.9\\
\rowcolor{mygray}
\textbf{Ties-Merging w/ \methodname} (Ours)  &69.8 &66.1 &87.3 &97.5 &86.7 &87.6 &98.5 &71.6 &83.1\\
\rowcolor{mygray}
\textbf{TW AdaMerging w/ \methodname} (Ours)  &63.9 &57.6 &84.2 &98.2 &87.6 & 92.7 &98.0 &66.8 &81.1\\
\rowcolor{mygray}
\textbf{AdaMerging w/ \methodname} (Ours)  &69.8&71.0 &88.9 &98.1 &91.7 &96.5 &98.8 &73.6 &86.1 \\
\rowcolor{mygray}
\textbf{AdaMerging w/ \methodname$^\dagger$} (Ours)  &71.2 &72.0 &92.3 &99.0 &92.2 &97.9 &99.0 &76.1 &87.5\\
\bottomrule
\multicolumn{10}{l}{{\footnotesize $^\dagger$ means that the rank (i.e., Eq.~\ref{eq:surgery}) of the surgery module is set to 64, and the other defaults are 16.}}
\end{tabular}
}
\vspace{-15pt}
\end{table*}

\subsection{Discussion}
\label{subsubsec:discussion}

\textbf{What is the positioning of ``representation surgery'' within the domain of model merging?}
The representation surgery proposed in this paper is orthogonal to existing model merging techniques in the following two aspects:
\begin{itemize}[noitemsep,topsep=0pt,parsep=0pt,partopsep=0pt,leftmargin=*]
    \item ``Seek common v.s. reserve differences'': All previous model merging methods focus on how to merge information shared by multiple tasks, that is, a process of \textit{``seeking common"}. On the contrary, the representation surgery proposed in this paper adds a task-specific module after the merged model to store task-private information, so it achieves \textit{\textbf{seeking common while reserving differences}}.
    \item ``Before-merging v.s. during-merging v.s. post-merging'': All previous model merging methods focus on how to create better merging conditions \textit{before merging} or mitigate conflicts \textit{during merging}, as discussed in Sec.~\ref{sec:relatedwork}. In contrast, our representation surgery focuses on how to solve the ``representation bias" problem \textit{\textbf{post-merging}}.
\end{itemize}
In summary, the representation surgery proposed in this paper is complementary to existing model merging schemes.

\textbf{Why does ``representation surgery" work?
}
As mentioned in Sec.~\ref{subsubsec:Objective}, the goal of the representation surgery is to reduce the discrepancy in the representation distributions between the merged and individual models.
Comparing the discrepancy in representation distributions in Fig.~\ref{fig:dis_wo_surgery} (w/o surgery) and Fig.~\ref{fig:dis_w_surgery} (w/ surgery), we can observe that the distributions of the merged and individual models in Fig.~\ref{fig:dis_w_surgery} are closer (i.e., higher overlap). 

Going one step further, we quantify the distance between the two distributions. Specifically, similar to the setting in Sec.~\ref{subsubsec:representation_bias}, we measure the $L_1$ distance between two sets (i.e., $\mathcal{Z}^{mtl}_{t}$ and $ \mathcal{Z}^{ind}_{t}$) of feature representations by Eq.~\ref{eq:l1distance}. As shown in Fig.~\ref{fig:l1_distance_vitb32}, we observe a significant reduction in the $L_1$ distance for the merged model with the proposed representation surgery (red column) compared to the merged model without representation surgery (blue column). For example, on the EuroSAT dataset, Task Arithmetic based model merging reduces the $L_1$ distance from 0.30 to 0.13 after representation surgery, a relative reduction of 56\%.
These pieces of evidence suggest that the proposed representation surgery helps alleviate the representation bias problem.

\section{Experiment}
\label{sec:experiment}

In this section, we describe our experimental setup and show performance comparisons. Due to the page limit, additional experimental setups and results are shown in the \textbf{Appendix}.

\begin{table*}[tbh!]
\centering
\caption{Multi-task performance when merging ViT-L/14 models on eight tasks.}
\label{tab:performance_vitlarge14} 
\resizebox{\linewidth}{!}{  
\begin{tabular}{l|cccccccc|cc}
\toprule
{Method}    &  {SUN397}  &  {Cars}  &  {RESISC45}  &  {EuroSAT}  &  {SVHN}  &  {GTSRB}  &  {MNIST}  &  {DTD}  & \textbf{Avg.} \\
\midrule
{Pretrained}  &  {66.8}  &  {77.7}  &  {71.0}  &  {59.9}  &  {58.4}  &  {50.5}  &  {76.3}  &  {55.3} & {64.5}   \\
{Individual}   &  82.3  &  92.4  &  97.4  &  100  &  98.1  &  99.2  &  99.7  &  84.1  & 94.2   \\
{Traditional MTL} &  80.8   &  90.6   &   96.3  & 96.3   & 97.6   & 99.1   &  99.6  &  84.4   & 93.5    \\
\midrule
{Weight Averaging}    &  72.1  &  81.6  &  82.6  &  91.9  &  78.2  &  70.7  &  97.1  &  62.8 & 79.6 \\
{Fisher Merging}~\citep{FisherMerging_NeurIPS2022}     &  69.2  &  88.6  &  87.5  &  93.5  &  80.6  &  74.8  &  93.3  &  70.0  & 82.2 \\
{RegMean}~\citep{RegMean_ICLR2023}    &  73.3  &  81.8  &  86.1  &  97.0  &  88.0  &  84.2  &  98.5  &  60.8  & 83.7 \\
\midrule
{Task Arithmetic}~\citep{TaskArithmetic_ICLR2023}  &73.9  &82.1 &86.6 &94.1  &87.9  &86.7  &98.9  &65.6   &84.5 \\
{Ties-Merging}~\citep{TiesMerging_NeurIPS2023}   &  76.5  &  85.0  &  89.3  &  95.7  &  90.3  &  83.3  &  99.0  &  68.8  & 86.0   \\
Concrete TA~\cite{tang2023concrete} &74.6 & 86.2 &89.0 & 96.7 &93.6 & 93.4  &99.1& 66.9  &87.4\\
Concrete AM~\cite{tang2023concrete} &77.8 &91.2 &92.1 &97.0 &94.4 &97.9 &99.0 &79.5 &91.1\\
{AdaMerging}~\cite{AdaMerging_Arxiv2023}   &79.0 &90.3 &90.8 & 96.2 &93.4 &98.0  &99.0  &79.9  &90.8 \\
\midrule
\rowcolor{mygray}
\textbf{Weight Averaging w/ \methodname} (Ours) &73.7 &83.9 &92.0 &98.4 &82.4 &86.3 &98.7 &71.9 &85.9 \\
\rowcolor{mygray}
\textbf{Task Arithmetic w/ \methodname} (Ours)  &75.7 &84.4 &93.1 &98.8 &91.3 &93.4 &99.1 &76.1 &89.0\\
\rowcolor{mygray}
\textbf{Ties-Merging w/ \methodname} (Ours)  &76.5 &85.9 &93.7 &99.2 &89.7 &92.0 &99.1 &78.1 &89.3\\
\rowcolor{mygray}
\textbf{AdaMerging w/ \methodname} (Ours)  &80.3 &90.8 &94.3 &98.2 &94.1 & 98.7 &99.2 &82.5 &92.3 \\
\bottomrule
\end{tabular}
}
\vspace{-15pt}
\end{table*}

\subsection{Experimental Setup}

\textbf{Datasets}. Following the setup of Task Arithmetic~\cite{TaskArithmetic_ICLR2023}, Ties-Merging~\cite{TiesMerging_NeurIPS2023} and AdaMerging~\cite{AdaMerging_Arxiv2023}, we treat the following eight datasets as eight tasks to perform model merging: SUN397~\citep{xiao2016sun}, Cars~\citep{krause20133d}, RESISC45~\citep{cheng2017remote}, EuroSAT~\citep{helber2019eurosat}, SVHN~\citep{yuval2011reading}, GTSRB~\citep{stallkamp2011german}, MNIST~\citep{lecun1998mnist}, DTD~\citep{cimpoi2014describing}.

\textbf{Baselines}. We use the following \textit{non-model merging methods} as reference baselines: Pretrained, Individual, Traditional MTL. Additionally, we compared the following \textit{model fusion methods}: Weight Averaging, Fisher Merging~\cite{FisherMerging_NeurIPS2022}, RegMean~\cite{RegMean_ICLR2023}, Task Arithmetic~\cite{TaskArithmetic_ICLR2023}, Ties-Merging~\cite{TiesMerging_NeurIPS2023}, Concrete TA~\cite{tang2023concrete}, Concrete AM~\cite{tang2023concrete}, AdaMerging~\cite{AdaMerging_Arxiv2023}.

\textbf{Architectures}. Following Task Arithmetic~\cite{TaskArithmetic_ICLR2023} and AdaMerging~\cite{AdaMerging_Arxiv2023}, we merge ViT-type architectures~\cite{Vit_ICLR2021}, including three different scale architectures of ViT-B/32, ViT-B/16 and ViT-L/14 from CLIP's~\cite{CLIP_ICML2021} visual encoder.

\subsection{Performance}

Due to page limitations, we leave the results of ViT-B/16 in the Appendix~\ref{sec:performance_appendix}. The results on the ViT-B/32 and ViT-L/14 architectures are shown in Tab.~\ref{tab:performance_vitbase32} and Tab.~\ref{tab:performance_vitlarge14}. From these two tables, we have the following observations: 
(i) For the \textit{\textbf{non-model merging baselines}}, the Pretrained model performed the worst, and the Individual model and Traditional MTL achieved optimal and suboptimal performance, respectively. The reason why a pre-trained model is bad is that it does not utilize any task-relevant information. In contrast, Traditional MTL uses data from all tasks to train the model together, significantly improving efficiency. However, Traditional MTL may suffer from negative transfer~\cite{negative_transfer_survey_2022} problems and, therefore, is not as effective as individual models.
(ii) For \textit{\textbf{model merging baselines}}, Weight Averaging is the simplest method. It directly averages multiple model parameters. Naturally, its performance is also the worst. Fisher Merging and RegMean calculate the importance of parameters/models during merging, and there are obvious hints in terms of performance compared to Weight Averaging. In addition, Concrete TA, Concrete AM and Ties-Merging remove some neurons when merging models, thereby effectively alleviating the parameter conflict problem during merging, and the final performance is better than baselines such as Task Arithmetic. TW AdaMerging and AdaMerging automatically learn task-wise/layer-wise merging coefficients on the test set in an unsupervised manner and achieve good results. However, as our analysis in Sec.~\ref{subsec:analysis} shows, these model merging methods still suffer from the problem of ``representation bias''.
(iii) Our proposed \textit{\textbf{representation surgery}} helps alleviate the representation bias problem, and it is orthogonal to existing model merging schemes. When the proposed representation surgery scheme is used on Weight Averaging, Task Arithmetic, Ties-Merging, and AdaMerging, their performance has been greatly improved. For example, on ViT-B/32, the accuracy of Task Arithmetic without and with the proposed representation surgery is 69.1\% and 80.9\%, respectively. On the more advanced AdaMerging, the accuracy has also increased from 80.1\%  to 87.5\% , which is very close to the 88.9\%  of Traditional MTL. On ViT-L/14, AdaMerging achieved an accuracy of 92.3\% after representation surgery, which is also very close to the 93.5\% of Traditional MTL.

\begin{figure}[t]
\vspace{-5pt}
    \centering 
    \includegraphics[width=0.44\textwidth]{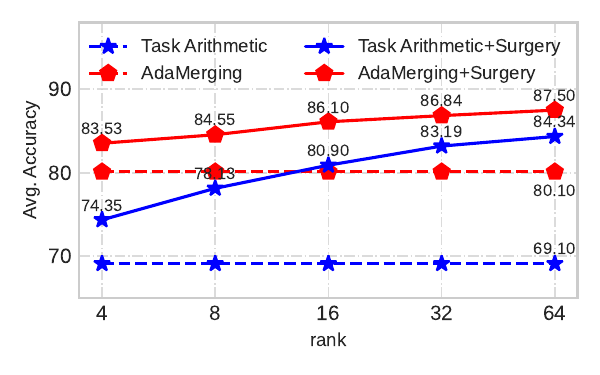}
    \vspace{-20pt}
    \caption{The average accuracy changes corresponding to different ranks in the surgery module under ViT-B/32 architecture.}
\label{fig:rank} 
\vspace{-15pt}
\end{figure}

\subsection{Additional Results and Analysis}
\label{subsec:analysis}

\textbf{Rank in Surgery Module}. As shown in Sec.~\ref{subsubsec:discussion}, the representation surgery module can be regarded as used to save the task-private information of each task. Therefore, the capacity of this module has an impact on the accuracy of model merging. In this section, we try different ranks (i.e., $r \in \{4, 8, 16, 32, 64\}$ in Eq.~\ref{eq:surgery}) and observe the accuracy changes. As shown in Fig.~\ref{fig:rank} and Tab.~\ref{tab:rank_vitbase32_appendix} in Appendix~\ref{sec:performance_appendix}, we consistently observed that as the rank increases, the accuracy of model merging also improves. For example, on AdaMerging, when the rank increases from 16 to 64, the average accuracy improves from 86.1\% to 87.5\%.

\begin{figure}[t]
    \centering 
     \includegraphics[width=0.43\textwidth]{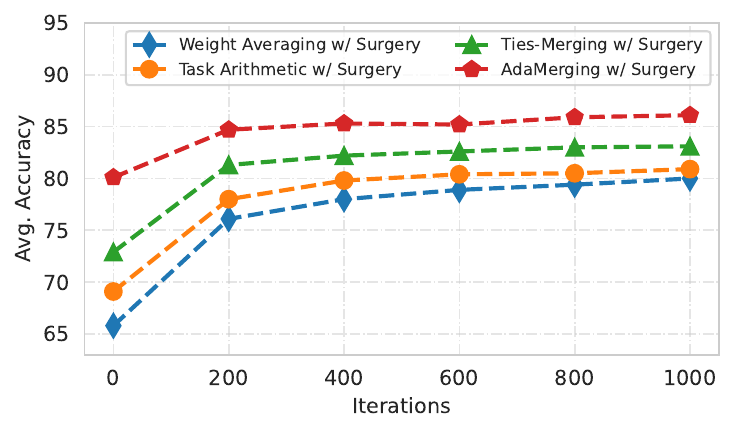}
    \vspace{-18pt}
    \caption{The average accuracy of model merging changes with the number of iterations on ViT-B/32.}  
\label{fig:iter_acc} 
\vspace{-16pt}
\end{figure}

\textbf{Training Step v.s. Avg. Accuracy}. 
In Fig.~\ref{fig:iter_acc}, we show how the average accuracy of the merged model changes with the number of iterations of the representation surgery on the ViT-B/32. We observe that in the early stages of training (e.g., the first 200 iterations), the accuracy of the merged model improves rapidly. As the iterations continue, the accuracy improvement gradually flattens out.

\section{Conclusion and Future Work}
\label{sec:conclusion}

This paper first studies that the ``representation bias'' problem exists widely in model merging, and it exists across tasks, across model merging methods, and across architectures. Next, we propose a ``representation surgery'' scheme to alleviate the representation bias problem by reducing the difference between the merged model and the individual models. Finally, we have verified through a large number of experiments that existing model merging methods can effectively improve the performance of model merging using the proposed representation surgery.
In the future, we plan to apply the proposed representation surgery to more model merging schemes and explore model merging from different architectures or initializations.

\section*{Acknowledgments}
Li Shen is supported by STI 2030—Major Projects (No. 2021ZD0201405). Enneng Yang and Guibing Guo are supported by the National Natural Science Foundation of China under Grant No. 62032013, the Science and technology projects in Liaoning Province (No. 2023JH3/10200005), and the Fundamental Research Funds for the Central Universities under Grant No. N2317002. Xiaojun Chen is supported by NSFC under Grant no. 92270122; and in part by Guangdong Provincial Natural Science Foundation under grant no. 2023A1515012584; and in part by the Shenzhen Research Foundation for Basic Research, China, under Grant JCYJ20210324093000002. Dacheng Tao's research is partially supported by NTU RSR and Start Up Grants.

\section*{Impact Statement}
Model merging based MTL provides an orthogonal perspective to perform multi-task learning. This paper discusses the common problem of representation bias in model merging based MTL methods and proposes representation surgery to alleviate the representation bias problem.
This work has no ethical aspects as well as negative social consequences.

\bibliography{ref_short}
\bibliographystyle{icml2024}

\newpage
\appendix
\onecolumn
\appendix

\textbf{Appendix Overview}.
The main contents of this appendix are as follows:
\begin{itemize}[noitemsep,topsep=0pt,parsep=0pt,partopsep=0pt]
    \item In Appendix~\ref{sec:experimentalsetting_appendix}, we describe the datasets, baselines, and training details in detail.
    \item In Appendix~\ref{sec:performance_appendix},  we show some experimental results and experimental analyses that are deleted due to the page limit of the main text.
\end{itemize}

\section{Experimental Setting}
\label{sec:experimentalsetting_appendix}

\subsection{Datasets}
Following Task Arithmetic~\cite{TaskArithmetic_ICLR2023}, Ties-Merging~\cite{TiesMerging_NeurIPS2023} and AdaMerging~\cite{AdaMerging_Arxiv2023}, we merge the models trained on the following eight datasets.
\begin{itemize}[noitemsep,topsep=0pt,parsep=0pt,partopsep=0pt]
    \item \textbf{SUN397}~\citep{xiao2016sun}: The database is a benchmark dataset for Scene Understanding (SUN) and contains a total of 108,753 images from 397 classes, where each class contains a different number of images, but each class has at least 100 images.
    \item \textbf{Cars}~\citep{krause20133d}: Stanford Cars is a dataset used for fine-grained recognition in the field of computer vision. It contains images of 196 car classes with a total of 16,185 images. The images of each class are divided roughly 1:1 into training set and test set. 
    \item \textbf{RESISC45}~\citep{cheng2017remote}: The RESISC45 dataset is a publicly available benchmark for scene classification in remote sensing images. It contains 45 scene classes and each class contains 700 images (each image resolution is 256$\times$256), for a total of about 31,500 images.
    \item \textbf{EuroSAT}~\citep{helber2019eurosat}: EuroSAT is a Sentinel-2-based satellite image dataset primarily used to classify land use in geospatial imagery and contains 27,000 labeled and geo-referenced images in 10 classes.
    \item \textbf{SVHN}~\citep{yuval2011reading}: SVHN is a real image dataset containing 10 classes of color house number images, SVHN is very similar in style to MNIST~\citep{lecun1998mnist} (handwritten digits in grayscale images), but it contains a larger number of images (more than 600,000 digital images).
    \item \textbf{GTSRB}~\citep{stallkamp2011german}: The German Traffic Sign Recognition Benchmark (GTSRB) contains images with different lighting conditions and rich backgrounds. These images are classified into 43 classes of traffic signs, totaling more than 50,000 images.
    \item \textbf{MNIST}~\citep{lecun1998mnist}: MNIST is a large database of handwritten digits in 10 classes that is one of the most famous datasets in machine learning. It contains 60,000 training images and 10,000 test images, each of which is 28x28 pixels.
    \item \textbf{DTD}~\citep{cimpoi2014describing}: The Describable Textures Dataset (DTD) contains 5,640 real labeled texture images, divided into 47 classes, each with about 120 images (all between 300$\times$300 and 640$\times$640 pixels).
\end{itemize}

\subsection{Baselines}
The comparison methods in all our experiments are divided into three categories in total: non-model merging methods, model merging methods, and our methods. Specific information is as follows:

(i) \textit{\textbf{Non-model merging methods}}:
\begin{itemize}[noitemsep,topsep=0pt,parsep=0pt,partopsep=0pt]
    \item \textbf{Pretrained} directly uses the pre-trained model to predict multiple tasks. Since it does not utilize any downstream task-related information for model training, its performance on these downstream tasks is usually very poor.
    \item \textbf{Individual} makes predictions using models that are fine-tuned independently for each task. It is usually the best performance because it has no interference between tasks. However, it requires maintaining a copy of the model parameters for each downstream task, which can be prohibitive in terms of memory cost.
    \item \textbf{Traditional MTL} mixes training data from multiple tasks to fine-tune a pre-trained model (i.e., a hard-parameter sharing network). Due to potential interference between tasks, it often suffers from negative transfer~\cite{negative_transfer_survey_2022} problems, that is, the predictive performance on a single task is not as good as that of individual models. However, it is very attractive in terms of parameters and computational efficiency.
\end{itemize}

(ii) \textit{\textbf{Model merging methods}}:
\begin{itemize}[noitemsep,topsep=0pt,parsep=0pt,partopsep=0pt]
    \item \textbf{Weight Averaging} is the most direct and simple model merging method. It directly averages the model parameters trained on multiple tasks into one model to perform multi-task learning. 
    \item \textbf{Fisher Merging}~\cite{FisherMerging_NeurIPS2022} measures the importance of each parameter through the Fisher information matrix~\cite{fisher1922mathematical}, thus merging model parameters based on this importance.
    \item \textbf{RegMean}~\cite{RegMean_ICLR2023} adjusts the weights and forms linear combinations of rows in weight matrices using statistical information derived from training data.
    \item \textbf{Task Arithmetic}~\cite{TaskArithmetic_ICLR2023} defines the concept of ``task vector'', which takes the fine-tuned model parameters minus the pre-trained model parameters as a task vector, and then combines multiple task vectors and adds them to the pre-trained model to perform multi-task learning.
    \item \textbf{Ties-Merging}~\cite{TiesMerging_NeurIPS2023} adds three steps, TRIM, ELECT SIGN and MERGE, based on Task Arithmetic~\cite{TaskArithmetic_ICLR2023}. These steps delete unimportant parameters in the task vector and correct the sign conflict problem of parameters, thereby easing the interference when the final task vector is merged. 
    \item \textbf{Concrete TA}~\cite{tang2023concrete} merges models in a parameter subspace shared between tasks, where the subspace is a learnable mask matrix, and Task Arithmetic~\cite{TaskArithmetic_ICLR2023} is used when merging models in the subspace.
    \item \textbf{Concrete AM}~\cite{tang2023concrete} also merges models in subspace, and uses AdaMerging~\cite{AdaMerging_Arxiv2023} to learn model merging coefficients during merging.
    \item \textbf{TW AdaMerging}~\cite{AdaMerging_Arxiv2023} uses an unsupervised test set to adaptively learn the merging coefficient of each task vector in Task Arithmetic~\cite{TaskArithmetic_ICLR2023}.
    \item \textbf{AdaMerging}~\cite{AdaMerging_Arxiv2023} uses an unlabeled test set to adaptively learn the merging coefficients of each layer in each task vector in Task Arithmetic~\cite{TaskArithmetic_ICLR2023}.
\end{itemize}

(iii) \textit{\textbf{Our methods}}:

Note that our representation surgery scheme is orthogonal to existing model merging schemes and can be seamlessly integrated into arbitrary model merging schemes. In this paper, we choose four representative model merging methods for experiments.
\begin{itemize}[noitemsep,topsep=0pt,parsep=0pt,partopsep=0pt]
    \item \textbf{Weight Averaging w/ \methodname (Ours)}: The representation surgery scheme proposed in this paper is performed on the model merged using the Weight Averaging scheme.
    \item \textbf{Task Arithmetic w/ \methodname (Ours)}: Based on Task Arithmetic~\cite{TaskArithmetic_ICLR2023}, the representation surgery scheme proposed in this paper is adopted.
    \item \textbf{Ties-Merging w/ \methodname (Ours)}: The representation surgery scheme, as suggested in this paper, is executed on the model created through the Ties-Merging~\cite{TiesMerging_NeurIPS2023} scheme. 
    \item \textbf{AdaMerging w/ \methodname (Ours)}: The proposed representation surgery scheme is applied on AdaMerging~\cite{AdaMerging_Arxiv2023}, an advanced model merging method.
\end{itemize}

\subsection{Discussion of Related Work}
By reading related work recommended by the anonymous reviewer, we find that this work has some technical connections with KD4MTL~\cite{kd4mtl2020}, but also has the following essential differences:
\begin{itemize}[noitemsep,topsep=0pt,parsep=0pt,partopsep=0pt]
    \item \textbf{Different learning paradigms}: KD4MTL belongs to the traditional MTL paradigm, which trains an MTL from scratch (i.e., Learn From Data) on raw data from multiple tasks. However, our work directly combines multiple independently trained models to complete MTL, and provides a new scheme for MTL (i.e., Learn From Model), which is orthogonal to traditional MTL. In addition, the paradigm of model merging based MTL effectively reduces the data management cost and data privacy issues of training data.
    \item \textbf{Different goals}: KD4MTL uses the trained model as an additional regularization term to alleviate the imbalance loss optimization problem in traditional MTL, which is caused by differences in task difficulty and loss magnitude. On the other hand, our work focuses on mitigating the representation bias problem caused by model merging, which is caused by parameter interference.
    \item \textbf{Different learning resources}: Our approach only requires some individually trained models to be merged, which is a more practical scenario. In contrast, KD4MTL requires both individually trained models and raw data to be trained from scratch, which is more expensive.
\end{itemize}
It should be noted that the traditional MTL paradigm, including KD4MTL~\cite{kd4mtl2020} and Naive MTL, is an \textit{upper bound} of the model merging based MTL (as shown in Tab.~\ref{tab:bert_performance_appendix}). In this paper, we focus on comparing various model merging based MTL methods.

\subsection{Implementation Details}
Our proposed representation surgery module contains a small number of trainable parameters (i.e., $\small \mathbf{W}_{down}$ and $\small \mathbf{W}_{up}$ in Eq.~\ref{eq:surgery}). In this paper, we do not require arbitrarily labeled training data. Instead, we use unlabeled test data and individual models to construct self-supervised training signals to update these parameters. Specifically, we use the Adam optimizer~\cite{adam_2014} to update these parameters with a learning rate of $1e\!-\!3$ and a momentum of (0.9, 0.999). We update for 1,000 iterations with a batch size of 16. In addition, we set the rank (i.e., $r$ in Eq.~\ref{eq:surgery}) of the surgery module to 16 by default, and we also tried values such as \{4, 8, 16, 32, 64\} in the experimental analysis. Finally, we report the accuracy on each task, as well as the average accuracy (i.e., {\textbf{Avg.}}) on the eight tasks.

\section{Experimental Results and Analysis}
\label{sec:performance_appendix}

\subsection{Performance in Computer Vision Domain}

\textbf{Performance on ViT-B/16}.
Tab.~\ref{tab:performance_vitbase16_appendix} shows the results of various model merging methods in the ViT-B/16 architecture. We can observe that when using the proposed representation surgery module on Weight Averaging, Task Arithmetic, Ties-Merging and AdaMerging, the performance of all methods has been significantly improved. For example, on Weight Averaging, the performance without representation surgery is 70.7\%, but with the use of surgery, the performance is improved to 82.6\%. Similarly, on AdaMerging, with the use of surgery, the performance is improved from 84.9\% to 88.8\%.

\begin{table*}[tbh!]
\vspace{-5pt}
\centering
\caption{Multi-task performance when merging ViT-B/16 models on eight tasks.}
\label{tab:performance_vitbase16_appendix} 
\resizebox{\linewidth}{!}{  
\begin{tabular}{l|cccccccc|cc}
\toprule
{Method}   &  {SUN397}  &  {Cars}  &  {RESISC45}  &  {EuroSAT}  &  {SVHN}  &  {GTSRB}  &  {MNIST}  &  {DTD}  & \textbf{Avg.} \\
\midrule
Pretrained  &63.8 &64.6 &65.7 &54.5 & 52.0&43.3 & 51.7&45.1 &55.0\\
Individual  &81.8 &86.8 &96.9 &99.7 &97.8 &99.1 &99.7 &82.0 &92.9\\
\midrule
Weight Averaging  &67.7 &70.0 &75.3 &79.5 &74.9 &60.1 &94.4 &43.8 &70.7\\
Fisher-Merging~\citep{FisherMerging_NeurIPS2022}  &68.5 &69.9 &75.2 &80.4 &73.2 &61.2 &94.5 &50.7 &71.7\\
RegMean~\citep{RegMean_ICLR2023} &69.1 &71.6 &77.6 &88.8 &83.7 &70.2 &96.9 &54.6 &76.6\\
 \midrule
{Task Arithmetic}~\citep{TaskArithmetic_ICLR2023} &61.1 &65.9 &74.0 &76.2 &88.0 &73.9 &98.4  &53.0 &73.8 \\
{Ties-Merging}~\citep{TiesMerging_NeurIPS2023} &69.1 &72.5 &80.5 &84.0 &85.0 &71.5 &98.1  &54.9 &77.0 \\
{AdaMerging}~\cite{AdaMerging_Arxiv2023})  &70.2 &80.7 &81.6 &94.8 &91.6 &95.8 &98.5  & 66.2 &84.9 \\
\midrule
\rowcolor{mygray}
\textbf{Weight Averaging w/ \methodname} (Ours) &70.3 &72.4 &88.8 &97.6 &82.0 & 83.1 &98.1 &68.5 &82.6 \\
\rowcolor{mygray}
\textbf{Task Arithmetic w/ \methodname} (Ours)  &68.3 &72.3 &88.7 & 97.7 &91.0 &89.5 &98.9 &72.9 &84.9\\
\rowcolor{mygray}
\textbf{Ties-Merging w/ \methodname} (Ours)  &73.0 &76.2 &90.7 &98.1 & 89.7 &86.7 &98.7 &75.2 &86.0\\
\rowcolor{mygray}
\textbf{AdaMerging w/ \methodname} (Ours)  &73.6 &81.5 &90.4 &98.5 &93.2 &97.4 & 98.9 &77.0 &88.8\\
\bottomrule
\end{tabular}
}
\vspace{-5pt}
\end{table*}

\begin{figure*}[h]
\vspace{-10pt}
    \centering 
    \subfigure{
        \begin{minipage}[t]{0.48\linewidth}
        \includegraphics[width=1.0\textwidth]{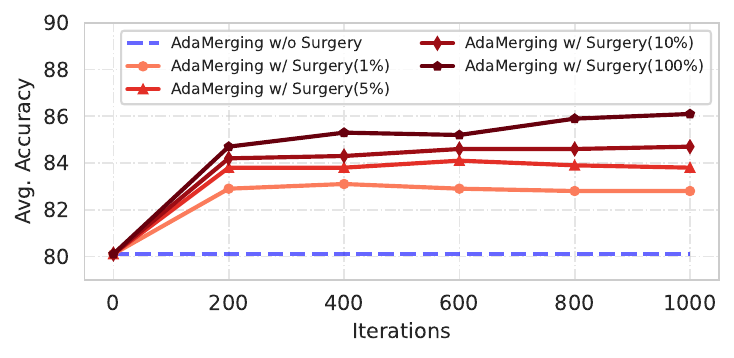}
        \vspace{-22pt}
        \begin{center}
            \text{\small \hspace{10pt} (a)}
         \end{center}
        \end{minipage}
    }
    \subfigure{
        \begin{minipage}[t]{0.48\linewidth}
        \includegraphics[width=1.0\textwidth]{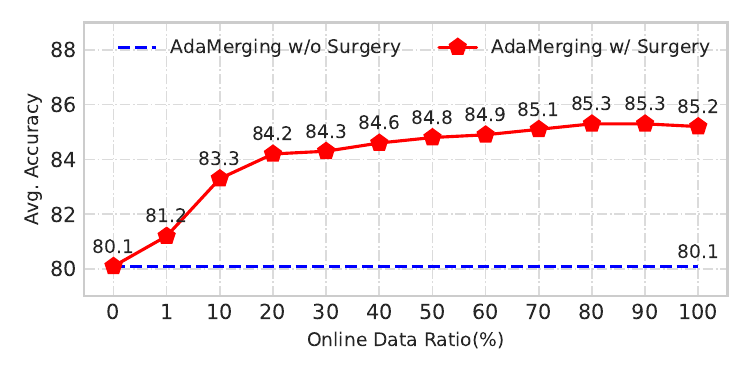}
        \vspace{-22pt}
        \begin{center}
            \text{\small \hspace{10pt} (b)}
         \end{center}
        \end{minipage}
    }
    \vspace{-12pt}
    \caption{
    (a) Performance variation of the ``representation surgery" scheme for different amounts of test data in the \textit{offline} scenario.
    (b) Performance change of the ``representation surgery'' scheme for different test data volumes in the \textit{online} scenario.}  
\label{fig:online_acc_appendix} 
\vspace{-5pt}
\end{figure*}

\textbf{Available Data Ratio}.
Our scheme relies on unlabeled test data to build self-supervision signals and thus update the ``representation surgery" module. As shown in Tab.~\ref{tab:testset_ratio_appendix} and Fig.~\ref{fig:online_acc_appendix}(a), we tested the performance of model merging using representation surgery when different ratios (e.g., 1\%, 5\%, 10\% or 100\%) of unlabeled test data are visible. We observe that the representation surgery is consistently effective for different amounts of data. For example, with only 1\% of visible data, the accuracy is 82.8\% using representation surgery, which is significantly higher than 80.1\% without surgery.  In addition, as the visible data increases, the gain obtained by representing the surgery clearly increases. For example, when 10\% of the data is visible, the average accuracy improves to 84.7\%.

\begin{table*}[h]
\vspace{-5pt}
\centering
\caption{Impact of the amount of available test data in ``representation surgery'' module on model performance.}
\label{tab:testset_ratio_appendix} 
\resizebox{\linewidth}{!}{  
\begin{tabular}{l|c|cccccccc|c}
\toprule
{Method}  & Available Test Set &  {SUN397}  &  {Cars}  &  {RESISC45}  &  {EuroSAT}  &  {SVHN}  &  {GTSRB}  &  {MNIST}  &  {DTD}  & \textbf{Avg.}  \\
\midrule
{AdaMerging}~\citep{AdaMerging_Arxiv2023}  & - &64.5 &68.1 &79.2 &93.8 &87.0 &91.9 &97.5  &59.1 &80.1   \\
\rowcolor{mygray}
\textbf{AdaMerging w/ \methodname} (Ours) & 1\% &68.6    &64.1   &85.3   &96.3   &90.9   & 95.9   & 98.4   &63.0   &82.8  \\
\rowcolor{mygray}
\textbf{AdaMerging w/ \methodname} (Ours) & 5\% &69.4    & 67.4  & 87.9  & 97.3  & 91.5  &  95.7  &  98.5  & 62.6  & 83.8 \\
\rowcolor{mygray}
\textbf{AdaMerging w/ \methodname} (Ours) & 10\% &69.7    &68.9   &88.0   &97.5   &91.5   &95.6    &98.5    & 67.8  & 84.7 \\
\rowcolor{mygray}
\textbf{AdaMerging w/ \methodname} (Ours) & 100\%  &69.8&71.0 &88.9 &98.1 &91.7 &96.5 &98.8 &73.6 &86.1 \\
\bottomrule
\end{tabular}
}
\vspace{-5pt}
\end{table*}

\textbf{Online Data Ratio}.
In a more realistic scenario, the test data may arrive \textit{online}, and receive only one sample at a time, where each sample is used to train the model only once rather than multiple times. As shown in Tab.~\ref{tab:online_testset_ratio_batchsize1_appendix} and Fig.~\ref{fig:online_acc_appendix}(b), we tested the performance of the proposed ``representation surgery" scheme in this scenario. We observe that when only a small amount of data is used to train the model once, there is also a performance gain compared to not using the representation surgery scheme. For example, with 10\% of the data, the average accuracy of the merged model with ``representation surgery'' is 83.3\%, while the average accuracy without ``representation surgery'' is 80.1\%. In addition, as the amount of online data increases, the average performance of using the ``representation surgery'' scheme improves significantly. For example, increasing from 10\% to 50\% improves the average performance from 83.3\% to 84.8\%.

% batch size = 1, single pass
\begin{table*}[h]
\vspace{-5pt}
\centering
\caption{Impact of the amount of \textbf{online test data} on model performance in ``representation surgery'' module.}
\label{tab:online_testset_ratio_batchsize1_appendix} 
\resizebox{\linewidth}{!}{  
\begin{tabular}{l|c|cccccccc|c}
\toprule
{Method}  & Available Test Set &  {SUN397}  &  {Cars}  &  {RESISC45}  &  {EuroSAT}  &  {SVHN}  &  {GTSRB}  &  {MNIST}  &  {DTD}  & \textbf{Avg.}  \\
\midrule
{AdaMerging}~\citep{AdaMerging_Arxiv2023}  & - &64.5 &68.1 &79.2 &93.8 &87.0 &91.9 &97.5  &59.1 &80.1 \\
\rowcolor{mygray}
\textbf{AdaMerging w/ \methodname} (Ours) & 1\%  &67.7   &68.0      &81.8    &91.2    &88.4   &94.5    &97.7   &60.5  &81.2 \\
\rowcolor{mygray}
\textbf{AdaMerging w/ \methodname} (Ours) & 10\% &69.2   &68.7      &85.2    &95.3    &90.0   &95.8    &98.2   &64.2  & 83.3     \\
\rowcolor{mygray}
\textbf{AdaMerging w/ \methodname} (Ours) & 20\% &69.5   &69.1      &86.7    &95.9    &91.1   &95.7    &98.2   &67.5  & 84.2     \\
\rowcolor{mygray}
\textbf{AdaMerging w/ \methodname} (Ours) & 30\% &69.8  &69.2       &87.0    &96.3    &91.1   &95.7    &98.4   &67.5  & 84.3     \\
\rowcolor{mygray}
\textbf{AdaMerging w/ \methodname} (Ours) & 40\% &69.6   &68.7      &87.5    &97.1    &91.4   &96.1    &98.5   &67.9  & 84.6      \\
\rowcolor{mygray}
\textbf{AdaMerging w/ \methodname} (Ours) & 50\% &69.5   &69.9      &87.9    &97.0    &91.3   &96.1    &98.5   &68.7  & 84.8     \\
\rowcolor{mygray}
\textbf{AdaMerging w/ \methodname} (Ours) & 60\% &69.6   &68.6      &87.9    &97.4    &91.6   &96.1    &98.5   &69.6  &84.9      \\
\rowcolor{mygray}
\textbf{AdaMerging w/ \methodname} (Ours) & 70\% &69.5   &69.5      &87.9    &97.1   &91.6   &96.2   &98.6   &71.0  &85.1      \\
\rowcolor{mygray}
\textbf{AdaMerging w/ \methodname} (Ours) & 80\% &69.7   &69.6      &88.4    &97.4    &91.4   &96.5    &98.5   &71.0  &85.3      \\
\rowcolor{mygray}
\textbf{AdaMerging w/ \methodname} (Ours) & 90\% &69.5   &69.7      &88.1    &97.6    &92.0   &96.4    &98.5   &70.6  &85.3      \\
\rowcolor{mygray}
\textbf{AdaMerging w/ \methodname} (Ours) & 100\% &69.4  &69.0      &88.2    &97.6    &92.0   &96.3    &98.7   &70.7  &85.2      \\
\bottomrule
\end{tabular}
}
\vspace{-5pt}
\end{table*}

\subsection{Performance in Natural Language Processing (NLP) Domain}

In this section, we demonstrate that the phenomenon of ``representation bias" also exists in the merged model in the NLP domain, and the performance of the merged model after applying the proposed ``representation surgery" is significantly improved. Specifically, as shown in Tab.~\ref{tab:bert_performance_appendix} and Tab.~\ref{tab:bert_bias_appendix}, we have the following observations: 
(i) Traditional MTL collects data from multiple tasks in advance and jointly trains an MTL model, which has better performance. In addition, more advanced MTL (e.g., KD4MTL~\cite{kd4mtl2020}) can further improve performance through some additional designs to alleviate the negative transfer problem in multi-task joint training. These learning-from-data MTL methods can be viewed as \textit{upper bounds} on model merging-based MTL methods.
(ii) Weight Averaging and Task Arithmetic suffer from the ``representation bias" problem (in Tab.~\ref{tab:bert_bias_appendix}), and the performance of the merged model has a significant gap compared with Traditional MTL or KD4MTL (in Tab.~\ref{tab:bert_performance_appendix}).
(iii) When the ``representation surgery" scheme proposed in this paper is used in the merged model, the representation bias is significantly alleviated (in Tab.~\ref{tab:bert_bias_appendix}), and the performance of the merged model is also significantly improved (in Tab.~\ref{tab:bert_performance_appendix}), very close to traditional MTL.

\begin{table*}[tbh!]
\small
\vspace{-5pt}
\centering
\caption{{Multi-task performance (higher better) when merging BERT models on five NLP tasks.}}
\label{tab:bert_performance_appendix} 
\resizebox{\linewidth}{!}{  
\begin{tabular}{l|c|ccccc|cccc}
\toprule
{Method}  &Learning Source&  {AG News}  &  {Yelp Sentiment}  &  {Amazon Sentiment}  &  {Yahoo Q\&A}   &  {DBPedia Wikipedia}  & \textbf{Avg.}  \\
\midrule
Traditional MTL  &Original Training Data & 90.6 & 59.1 & 55.6 & 71.3 & 98.5 & 75.0  \\
KD4MTL~\cite{kd4mtl2020} & Original Training Data
\& Trained Models&   91.6 & 59.2 & 57.0 & 71.2 & 98.7 & 75.5  \\
\midrule
Weight Averaging & Trained Models& 79.2 & 49.8 & 45.0 & 50.3 & 55.1 & 55.8 \\
\rowcolor{mygray}
\textbf{Weight Averaging w/ \methodname} (Ours)&Trained Models& 90.3 & 58.0 & 54.2 & 70.8 & 98.4& 74.3 \\
Task Arithmetic~\citep{TaskArithmetic_ICLR2023}  &Trained Models& 82.9 & 55.8 & 48.4 & 53.1 & 81.5 &  64.3  \\
\rowcolor{mygray}
\textbf{Task Arithmetic w/ \methodname} (Ours) &Trained Models& 89.8 & 58.4 & 55.4 & 70.3& 98.0& 74.4 \\
\bottomrule
\end{tabular}
}
\vspace{-5pt}
\end{table*}

\begin{table*}[tbh!]
\small
\vspace{-5pt}
\centering
\caption{{Representation bias (lower better) when merging BERT models on five NLP tasks.}}
\label{tab:bert_bias_appendix} 
\resizebox{\linewidth}{!}{  
\begin{tabular}{l|ccccc|ccccc}
\toprule
{Method}  &  {AG News}  &  {Yelp Sentiment}  &  {Amazon Sentiment}  &  {Yahoo Q\&A}   &  {DBPedia Wikipedia}  & \textbf{Avg.}  \\
\midrule
Weight Averaging  & 0.448 & 0.336 & 0.349 & 0.418 & 0.539 & 0.418 \\
\rowcolor{mygray}
\textbf{Weight Averaging w/ \methodname} (Ours) & 0.208 & 0.171 & 0.189 & 0.211 & 0.188 & 0.193  \\
Task Arithmetic~\citep{TaskArithmetic_ICLR2023}  & 0.373 & 0.305 & 0.362 & 0.378 & 0.395 & 0.362  \\
\rowcolor{mygray}
\textbf{Task Arithmetic w/ \methodname} (Ours) & 0.190 & 0.179 & 0.194 & 0.226 & 0.172 & 0.192  \\
\bottomrule
\end{tabular}
}
\vspace{-5pt}
\end{table*}

\subsection{Analysis}
\label{sebsec:analysis}

\textbf{Different Ranks}.
Tab.~\ref{tab:rank_vitbase32_appendix} shows the performance changes when the surgery module uses different ranks under the ViT-B/32 architecture. We have consistently observed that in both model merging methods, Task Arithmetic and AdaMerging, as the rank size increases, the performance of the merged model improves. For example, when the rank $r$ increases from 4 to 64, Task Arithmetic's performance improves from 74.3\% to 84.3\%, while AdaMerging's performance improves from 83.5\% to 87.5\%. This is due to the fact that as mentioned in Sec.~\ref{subsubsec:discussion}, the surgery module can be regarded as being used to accommodate task-private information. When the dimension of the rank is larger, more task-private information can be stored, so the performance is better.

\begin{table*}[tbh!]
\vspace{-5pt}
\centering
\caption{Multi-task performance on the ViT-B/32 model when different ranks in the representation surgery module.}
\label{tab:rank_vitbase32_appendix} 
\resizebox{\linewidth}{!}{  
\begin{tabular}{l|cccccccc|cc}
\toprule
{Method}  &  {SUN397}  &  {Cars}  &  {RESISC45}  &  {EuroSAT}  &  {SVHN}  &  {GTSRB}  &  {MNIST}  &  {DTD}  & \textbf{Avg.}  \\
\midrule
Task Arithmetic~\citep{TaskArithmetic_ICLR2023}   &55.2 &54.9 &66.7 &78.9 &80.2 & 69.7 &97.3 &50.4 & 69.1 \\
\rowcolor{mygray}
\textbf{Task Arithmetic w/ \methodname} ($r$=4) &62.6 &55.9 &76.7 &78.7 &83.4 &79.6 &97.7 &60.0 &74.3\\
\rowcolor{mygray}
\textbf{Task Arithmetic w/ \methodname} ($r$=8) &63.2 &58.5 &79.9 &93.9 &84.5 &82.1 &98.5 &64.2 &78.1\\
\rowcolor{mygray}
\textbf{Task Arithmetic w/ \methodname} ($r$=16)  &63.8 &59.9 &83.3 &97.9 &87.0 &87.0 &98.6 &69.4 &80.9\\
\rowcolor{mygray}
\textbf{Task Arithmetic w/ \methodname} ($r$=32) &64.7 &62.4 & 87.3 &98.3 &88.3 &92.8 &98.8 &72.5 &83.1\\
\rowcolor{mygray}
\textbf{Task Arithmetic w/ \methodname} ($r$=64) &65.6 &62.7 &89.8 &98.3 & 88.7 &95.6 &98.9 &74.7 &84.3\\
\midrule
{AdaMerging}~\cite{AdaMerging_Arxiv2023}   &64.5 &68.1 &79.2 &93.8 &87.0 &91.9 &97.5  &59.1 &80.1 \\
\rowcolor{mygray}
\textbf{AdaMerging w/ \methodname} ($r$=4) &68.7 &68.9 &85.5 &95.6 & 88.3 &95.6 &97.9 &67.4 &83.5 \\
\rowcolor{mygray}
\textbf{AdaMerging w/ \methodname} ($r$=8) &69.0 &68.1 & 87.0 & 97.4 &89.3 &95.6 & 98.4 &71.2 &84.5\\
\rowcolor{mygray}
\textbf{AdaMerging w/ \methodname} ($r$=16)  &69.8&71.0 &88.9 &98.1 &91.7 &96.5 &98.8 &73.6 &86.1 \\
\rowcolor{mygray}
\textbf{AdaMerging w/ \methodname} ($r$=32) &70.5 &70.9 &90.4 &98.6 &92.1 &97.5 &98.8 &75.4 &86.8 \\
\rowcolor{mygray}
\textbf{AdaMerging w/ \methodname} ($r$=64) &71.2 &72.0 &92.3 &99.0 &92.2 &97.9 &99.0 &76.1 &87.5\\
\bottomrule
\end{tabular}
}
\vspace{-5pt}
\end{table*}

\textbf{Different Loss Functions}.
We aim to minimize the gap between "representations after performing surgery" and "representations from individual models". Therefore, any distance function can be used, not limited to $L_1$ loss in Eq.~\ref{eq:surgery}.
As shown in the following Tab.~\ref{tab:lossfun_vitbase32_appendix}, we further test the effectiveness of our method under different loss functions: MSELoss, SmoothL1Loss, and Negative CosineSimilarity. The proposed surgery method is effective under all these loss functions, further demonstrating the robustness of our approach.

\begin{table*}[h]
\vspace{-5pt}
\centering
\caption{Multi-task performance on the ViT-B/32 model when different loss functions are used in the representation surgery module.}
\label{tab:lossfun_vitbase32_appendix} 
\resizebox{\linewidth}{!}{  
\begin{tabular}{l|cccccccc|cc}
\toprule
{Method}  &  {SUN397}  &  {Cars}  &  {RESISC45}  &  {EuroSAT}  &  {SVHN}  &  {GTSRB}  &  {MNIST}  &  {DTD}  & \textbf{Avg.}  \\
\midrule
Task Arithmetic~\citep{TaskArithmetic_ICLR2023}   &55.2 &54.9 &66.7 &78.9 &80.2 & 69.7 &97.3 &50.4 & 69.1 \\
\rowcolor{mygray}
\textbf{Task Arithmetic w/ \methodname} (L1 Loss)  &63.8 &59.9 &83.3 &97.9 &87.0 &87.0 &98.6 &69.4 &80.9\\
\rowcolor{mygray}
\textbf{Task Arithmetic w/ \methodname} (MSELoss)& 64.3 & 59.8 & 84.0 &97.8  &87.6  & 88.7 & 98.8   & 69.8 & 81.4 \\
\rowcolor{mygray}
\textbf{Task Arithmetic w/ \methodname} (SmoothL1Loss)& 64.1 & 59.7 & 84.1 & 97.9  & 88.1 &  89.7 & 98.8 &  70.6  & 81.6  \\
\rowcolor{mygray}
\textbf{Task Arithmetic w/ \methodname} (Negative CosineSimilarity) & 63.8  & 59.2 & 84.5  & 97.5 & 88.8 &  86.4 & 99.0 &  70.3 & 81.2 \\
\bottomrule
\end{tabular}
}
\vspace{-5pt}
\end{table*}

\textbf{Parameters Costs}.
As shown in Tab.~\ref{tab:parameter_cost_Appendix}, we counted the number of additional parameters introduced by the proposed ``representation surgery'' module. We observe that the number of parameters in the surgery module is insignificant compared to the number of parameters that need to be merged, which is usually around one in ten thousand. For example,  on ViT-B/32, the number of parameters to be merged is 907,589,640, while the number of parameters for the surgery module is only 131,072, and the latter only accounts for 0.014\% of the former.

\begin{table*}[h]
\vspace{-5pt}
\centering
\small
\caption{The parameter cost of the representation surgery module ($r=16$).}
\label{tab:parameter_cost_Appendix} 
\begin{tabular}{l|c|c|c}
\toprule
% ViT-B-32: [8, 113448705]
% ViT-B-16: [8, 111792129]
% ViT-L-14: [8, 342562049]
{Architectures}  & ViT-B/32  & ViT-B/16 &  ViT-L/14 \\
\midrule
{The total number of merged model parameters} &907,589,640  & 894,337,032   &    2,740,496,392  \\
{The total number of parameters in the representation surgery module} &131,072  &131,072  &196,608  \\
\midrule
Ratio &0.000144 &0.000146 &0.000071 \\
\bottomrule
\end{tabular}
% }
\vspace{-5pt}
\end{table*}

\begin{figure*}[h]
    \centering 
\vspace{-5pt}
    \subfigure{
        \begin{minipage}[t]{0.45\linewidth}
        \includegraphics[width=1.0\textwidth]{images/iter_acc_ViT-B-32.pdf}
        \vspace{-20pt}
        \begin{center}
            \text{\small \hspace{20pt} (a) ViT-B/32}
         \end{center}
        \end{minipage}
    }
    \subfigure{
        \begin{minipage}[t]{0.45\linewidth}
        \includegraphics[width=1.0\textwidth]{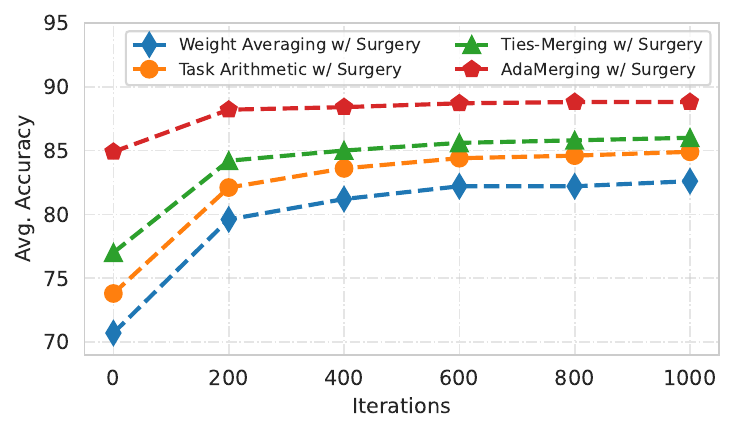}
        \vspace{-20pt}
        \begin{center}
            \text{\small \hspace{20pt}  (b) ViT-B/16}
         \end{center}
        \end{minipage}
    }
    \vspace{-5pt}
    \caption{The average accuracy of the merged model changes with the number of iterations on ViT-B/32 and ViT-B/16.}  
\label{fig:iter_acc_appendix} 
\vspace{-5pt}
\end{figure*}

\textbf{Training Step v.s. Avg. Accuracy}.
As shown in Fig.~\ref{fig:iter_acc_appendix}, we show how the average accuracy of the four model merging methods (i.e., Weight Averaging, Task Arithmetic, Ties-Merging and AdaMerging) using the proposed ``representation surgery" changes with the number of iterations on ViT-B/32 and ViT-B/16 architectures. We consistently observe that with only a small number of iterations (e.g., 200), the accuracy with representation surgery has a very significant improvement over the initial point (i.e., without surgery). This also indicates that the proposed ``representation surgery" scheme has the dual advantages of efficiency and effectiveness.

\subsection{Visualization}

\begin{figure*}[h]
    \centering 
    \includegraphics[width=.48\textwidth]{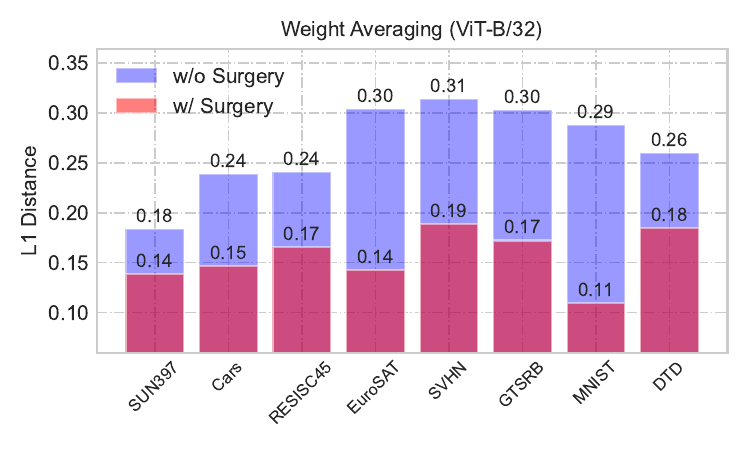}
    \includegraphics[width=.48\textwidth]{images/l1_distance/l1_distance_ViT-B-32_task_arithmetic.pdf}
    \includegraphics[width=.48\textwidth]{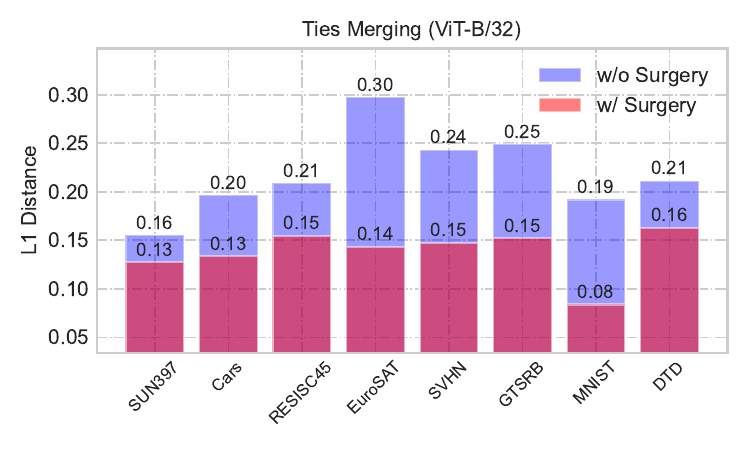}
    \includegraphics[width=.48\textwidth]{images/l1_distance/l1_distance_ViT-B-32_lw_adamerging.pdf}
    \vspace{-20pt}
    \caption{Visualization of the $L_1$ distance (or ``representation bias'' in Eq.~\ref{eq:l1distance}) of the representation of the merged model with and without representation surgery versus the individual model. All results are performed on \textbf{ViT-B/32} architecture.}  
\label{fig:l1_distance_vitb32_appendix} 
\vspace{-15pt}
\end{figure*}

\begin{figure*}[h]
    \centering 
    \includegraphics[width=.48\textwidth]{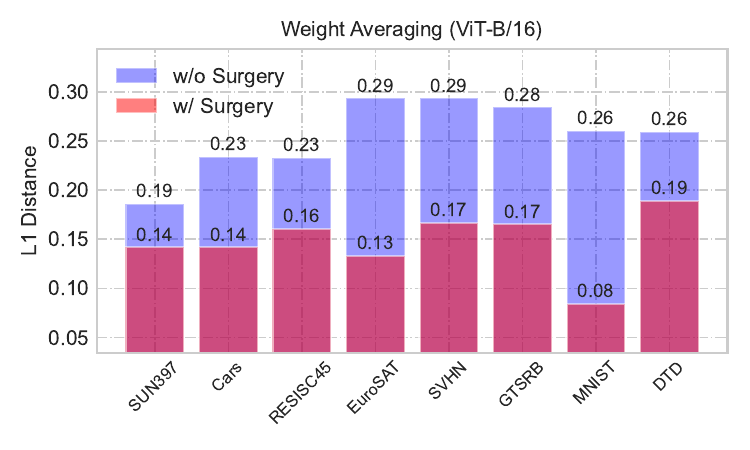}
    \includegraphics[width=.48\textwidth]{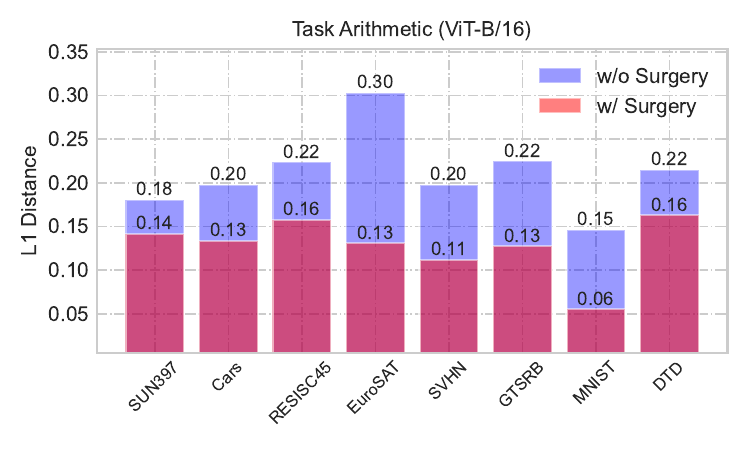}
    \includegraphics[width=.48\textwidth]{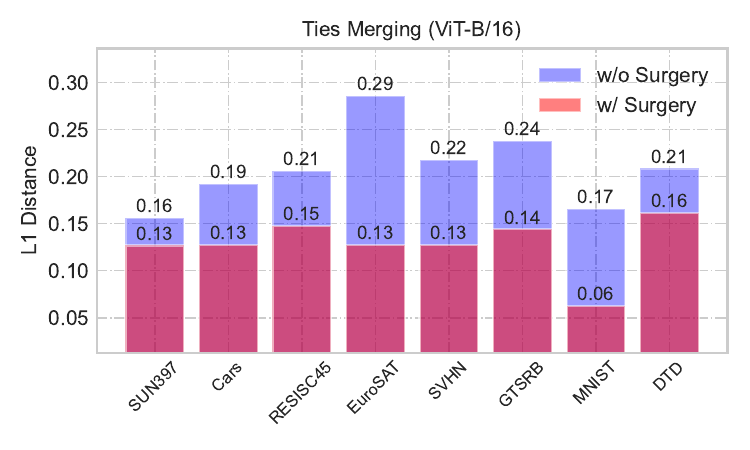}
    \includegraphics[width=.48\textwidth]{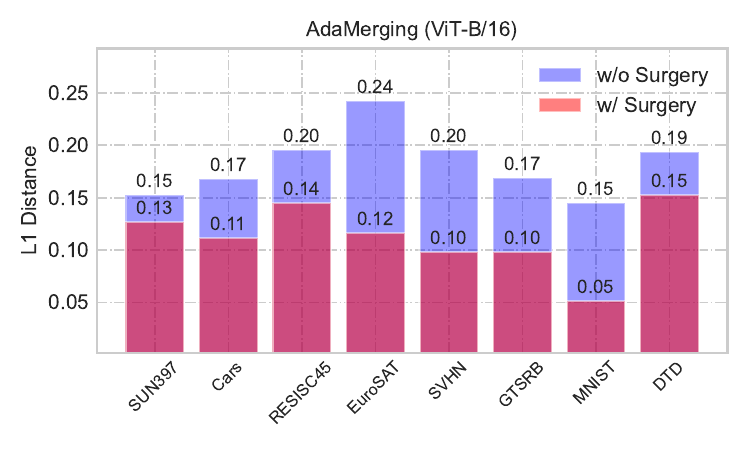}
    \vspace{-20pt}
     \caption{Visualization of the $L_1$ distance (or ``representation bias'' in Eq.~\ref{eq:l1distance}) of the representation of the merged model with and without representation surgery versus the individual model. All results are performed on \textbf{ViT-B/16} architecture.}  
\label{fig:l1_distance_vitb16_appendix} 
% \vspace{-15pt}
\end{figure*}

\begin{figure*}[h]
    \centering 
    \includegraphics[width=.48\textwidth]{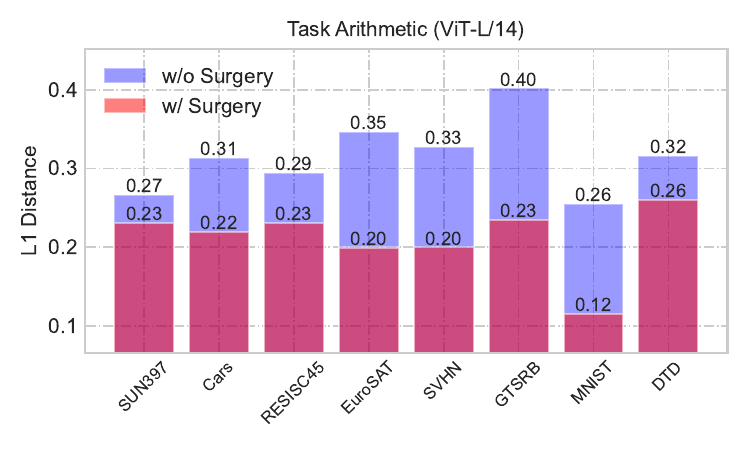}
    \includegraphics[width=.48\textwidth]{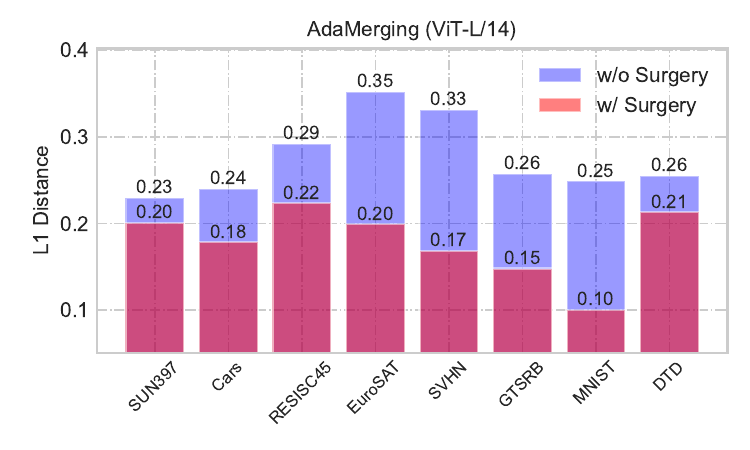}
    \vspace{-20pt}
     \caption{Visualization of the $L_1$ distance (or ``representation bias'' in Eq.~\ref{eq:l1distance}) of the representation of the merged model with and without representation surgery versus the individual model. All results are performed on \textbf{ViT-L/14} architecture.}  
\label{fig:l1_distance_vitl14_appendix} 
% \vspace{-15pt}
\end{figure*}

\textbf{Representation Bias or L1 Distance}. Fig.~\ref{fig:l1_distance_vitb32_appendix}, Fig.~\ref{fig:l1_distance_vitb16_appendix}, and Fig.~\ref{fig:l1_distance_vitl14_appendix} demonstrate the $L_1$ distances (i.e., Eq.~\ref{eq:l1distance}) of the feature representations extracted with and without surgery compared to the feature representations extracted by individual models on three architectures, ViT-B/32, ViT-B/16 and ViT-L/14, respectively. We have consistently observed that after using the representation surgery (red column), the $L_1$ distance is significantly reduced, which means that the representation surgery effectively alleviates the representation bias problem.

\textbf{Representation Distribution}. Tab.~\ref{tab:overall_distribution} shows that on three architectures of ViT-B/32, ViT-B/16 and ViT-L/14, representation distribution of the features extracted by the individual model versus the features extracted by the merged model with and without the proposed representation surgery for four model merging methods: Weight Averaging, Task Arithmetic~\cite{TaskArithmetic_ICLR2023}, Ties-Merging~\cite{TiesMerging_NeurIPS2023}, and AdaMerging~\cite{AdaMerging_Arxiv2023}.

\begin{table}[h]
\vspace{-5pt}
\centering
\caption{Visualization of the distribution of the four model merging methods performed and without performed representation surgery under the three architectures.}
\resizebox{\linewidth}{!}{  
\begin{tabular}{c|c|c|c|c}
\toprule
\multicolumn{1}{l|}{} & \multicolumn{1}{c|}{\textbf{Weight Averaging}} & \multicolumn{1}{c|}{\textbf{Task Arithmetic}} & \multicolumn{1}{c|}{\textbf{Ties-Merging}} & \multicolumn{1}{c}{\textbf{AdaMerging}} \\ \midrule
\textbf{ViT-B/32}      
&   w/o Surgery: Fig.~\ref{fig:distri_vitb32_avg_before}  v.s.   w/ Surgery: Fig.~\ref{fig:distri_vitb32_avg_after}                                        
&   w/o Surgery: Fig.~\ref{fig:distri_vitb32_tv_before}  v.s.   w/ Surgery: Fig.~\ref{fig:distri_vitb32_tv_after}     &  w/o Surgery: Fig.~\ref{fig:distri_vitb32_tiesmerging_before}  v.s.   w/ Surgery: Fig.~\ref{fig:distri_vitb32_tiesmerging_after}
&  w/o Surgery: Fig.~\ref{fig:distri_vitb32_ladamerging_before}  v.s.   w/ Surgery: Fig.~\ref{fig:distri_vitb32_ladamerging_after}                                                 
                                             \\ \midrule
\textbf{ViT-B/16}      
&  -                                        
&  -
&  -                                                 
&  w/o Surgery: Fig.~\ref{fig:distri_vitb16_lwadamerging_before}  v.s.   w/ Surgery: Fig.~\ref{fig:distri_vitb16_lwadamerging_after}  
\\ \midrule
\textbf{ViT-L/14}    
&  -                                        
&  -
&  -
&  w/o Surgery: Fig.~\ref{fig:distri_vitl14_lwadamerging_before}  v.s.   w/ Surgery: Fig.~\ref{fig:distri_vitl14_lwadamerging_after}  
\\ \bottomrule
\end{tabular}
}
\label{tab:overall_distribution}
\vspace{-5pt}
\end{table}

\begin{figure*}[h]
    \centering 
    \includegraphics[width=.24\textwidth]{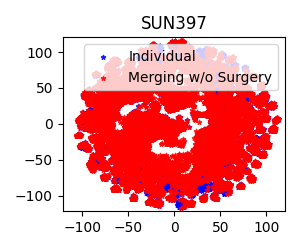}
    \includegraphics[width=.24\textwidth]{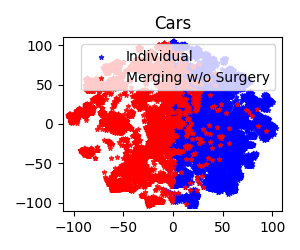}
    \includegraphics[width=.24\textwidth]{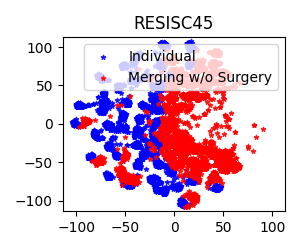}
    \includegraphics[width=.24\textwidth]{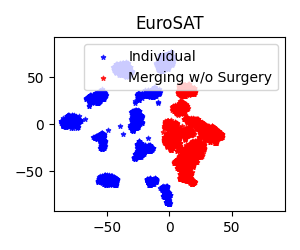}
    \includegraphics[width=.24\textwidth]{images/distribution/weight_averaging_ViT-B-32_SVHN_finetuned_Merging_wo_Surgery.png}
    \includegraphics[width=.24\textwidth]{images/distribution/weight_averaging_ViT-B-32_GTSRB_finetuned_Merging_wo_Surgery.png}
    \includegraphics[width=.24\textwidth]{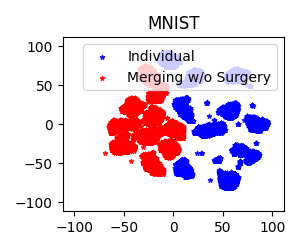}
    \includegraphics[width=.24\textwidth]{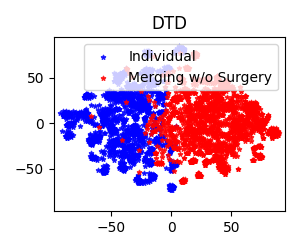}
    % \vspace{-10pt}
    \caption{Visualization of representation distribution of \textbf{Weight Averaging (w/o Surgery)} on \textbf{ViT-B/32} architecture.}  
\label{fig:distri_vitb32_avg_before} 
\end{figure*}

\begin{figure*}[h]
    \centering 
     \includegraphics[width=.24\textwidth]{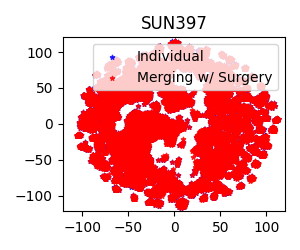}
    \includegraphics[width=.24\textwidth]{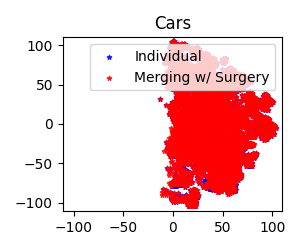}
    \includegraphics[width=.24\textwidth]{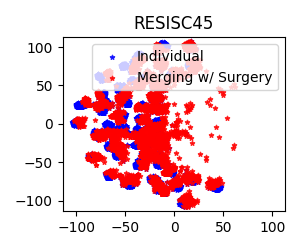}
    \includegraphics[width=.24\textwidth]{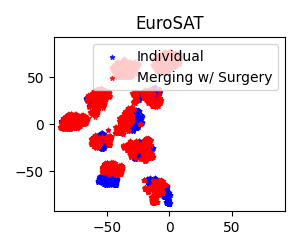}
    \includegraphics[width=.24\textwidth]{images/distribution/weight_averaging_ViT-B-32_SVHN_finetuned_Merging_w_Surgery.png}
    \includegraphics[width=.24\textwidth]{images/distribution/weight_averaging_ViT-B-32_GTSRB_finetuned_Merging_w_Surgery.png}
    \includegraphics[width=.24\textwidth]{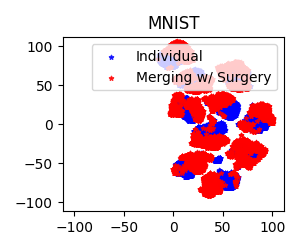}
    \includegraphics[width=.24\textwidth]{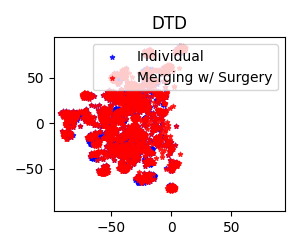}
    % \vspace{-10pt}
    \caption{Visualization of representation distribution of \textbf{Weight Averaging (w/ Surgery)} on \textbf{ViT-B/32} architecture.}  
\label{fig:distri_vitb32_avg_after} 
\end{figure*}

\begin{figure*}[h]
    \centering 
    \includegraphics[width=.24\textwidth]{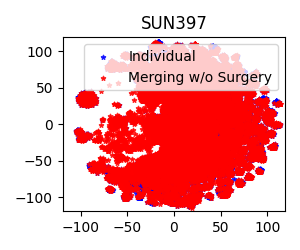}
    \includegraphics[width=.24\textwidth]{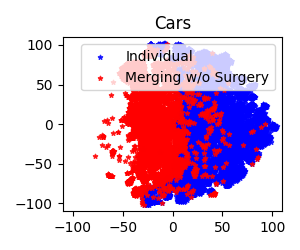}
    \includegraphics[width=.24\textwidth]{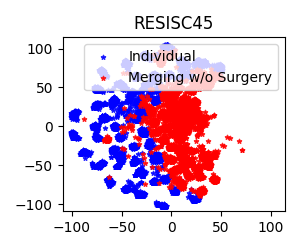}
    \includegraphics[width=.24\textwidth]{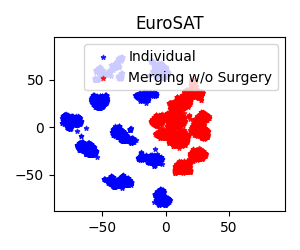}
    \includegraphics[width=.24\textwidth]{images/distribution/task_arithmetic_ViT-B-32_SVHN_finetuned_Merging_wo_Surgery.png}
    \includegraphics[width=.24\textwidth]{images/distribution/task_arithmetic_ViT-B-32_GTSRB_finetuned_Merging_wo_Surgery.png}
    \includegraphics[width=.24\textwidth]{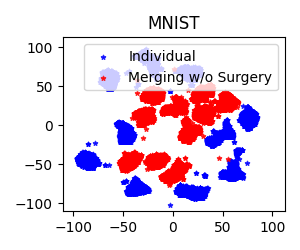}
    \includegraphics[width=.24\textwidth]{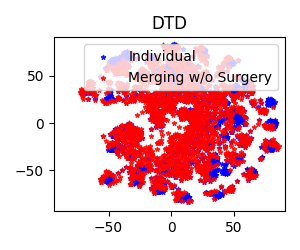}
    % \vspace{-10pt}
    \caption{Visualization of representation distribution of \textbf{Task Arithmetic (w/o Surgery)} on \textbf{ViT-B/32} architecture.}  
\label{fig:distri_vitb32_tv_before} 
\end{figure*}

\begin{figure*}[h]
    \centering 
     \includegraphics[width=.24\textwidth]{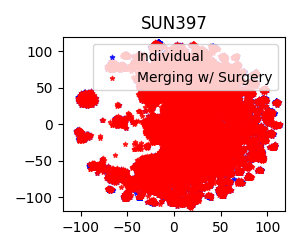}
    \includegraphics[width=.24\textwidth]{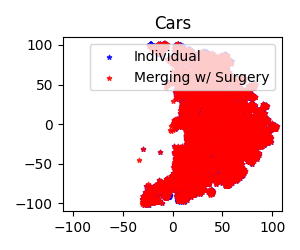}
    \includegraphics[width=.24\textwidth]{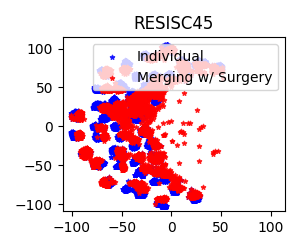}
    \includegraphics[width=.24\textwidth]{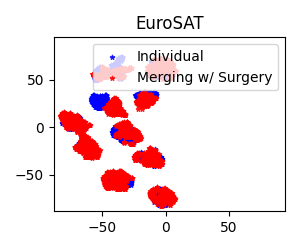}
    \includegraphics[width=.24\textwidth]{images/distribution/task_arithmetic_ViT-B-32_SVHN_finetuned_Merging_w_Surgery.png}
    \includegraphics[width=.24\textwidth]{images/distribution/task_arithmetic_ViT-B-32_GTSRB_finetuned_Merging_w_Surgery.png}
    \includegraphics[width=.24\textwidth]{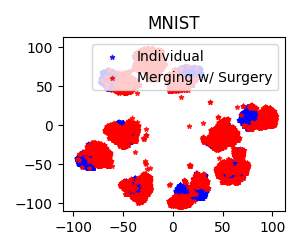}
    \includegraphics[width=.24\textwidth]{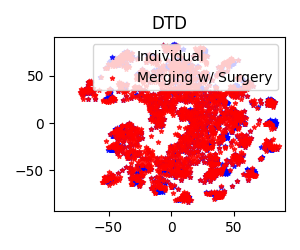}
    % \vspace{-10pt}
    \caption{Visualization of representation distribution of \textbf{Task Arithmetic (w/ Surgery)} on \textbf{ViT-B/32} architecture.}
\label{fig:distri_vitb32_tv_after} 
\end{figure*}

\begin{figure*}[h]
    \centering 
    \includegraphics[width=.24\textwidth]{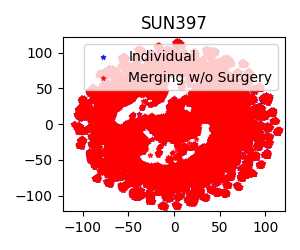}
    \includegraphics[width=.24\textwidth]{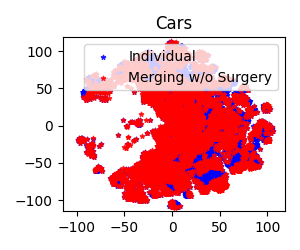}
    \includegraphics[width=.24\textwidth]{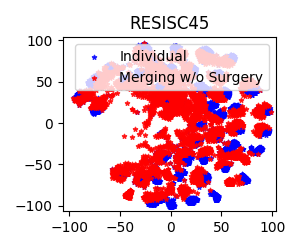}
    \includegraphics[width=.24\textwidth]{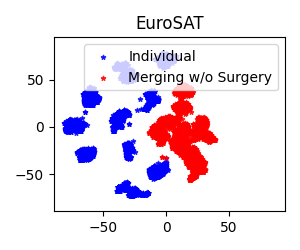}
    \includegraphics[width=.24\textwidth]{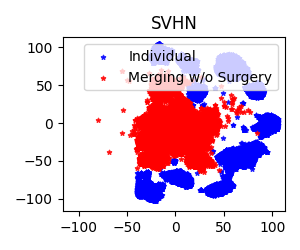}
    \includegraphics[width=.24\textwidth]{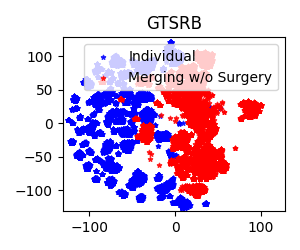}
    \includegraphics[width=.24\textwidth]{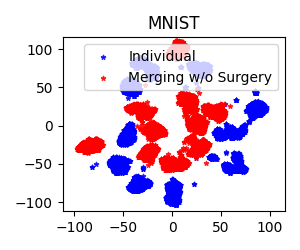}
    \includegraphics[width=.24\textwidth]{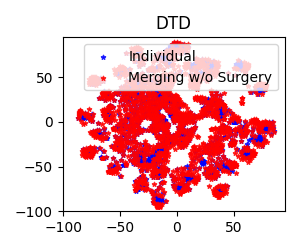}
    % \vspace{-10pt}
    \caption{Visualization of representation distribution of \textbf{Ties-Merging (w/o Surgery)} on \textbf{ViT-B/32} architecture.}
\label{fig:distri_vitb32_tiesmerging_before} 
\end{figure*}

\begin{figure*}[h]
    \centering 
     \includegraphics[width=.24\textwidth]{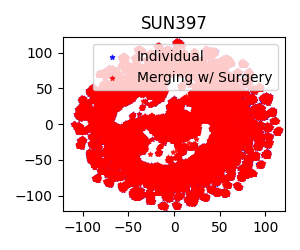}
    \includegraphics[width=.24\textwidth]{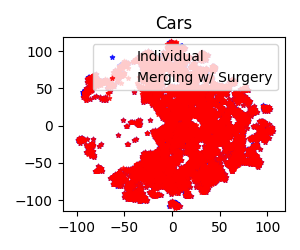}
    \includegraphics[width=.24\textwidth]{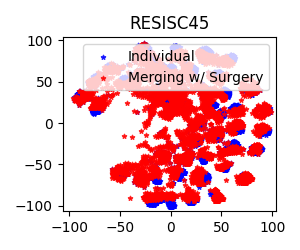}
    \includegraphics[width=.24\textwidth]{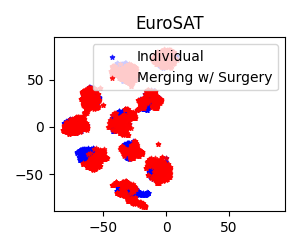}
    \includegraphics[width=.24\textwidth]{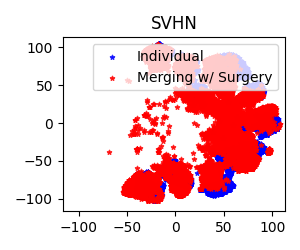}
    \includegraphics[width=.24\textwidth]{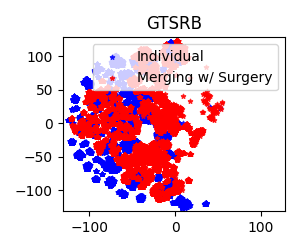}
    \includegraphics[width=.24\textwidth]{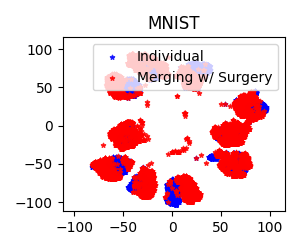}
    \includegraphics[width=.24\textwidth]{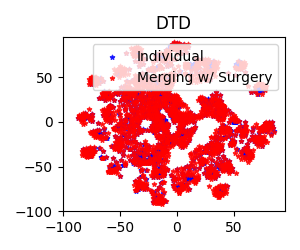}
    % \vspace{-10pt}
     \caption{Visualization of representation distribution of \textbf{Ties-Merging (w/ Surgery)} on \textbf{ViT-B/32} architecture.}
\label{fig:distri_vitb32_tiesmerging_after} 
\end{figure*}

\begin{figure*}[h]
    \centering 
    \includegraphics[width=.24\textwidth]{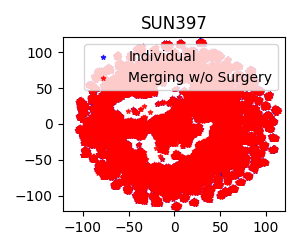}
    \includegraphics[width=.24\textwidth]{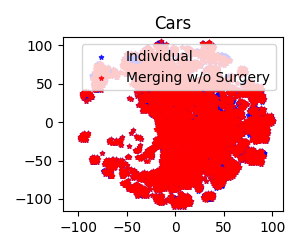}
    \includegraphics[width=.24\textwidth]{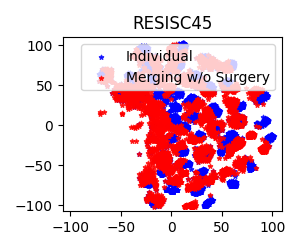}
    \includegraphics[width=.24\textwidth]{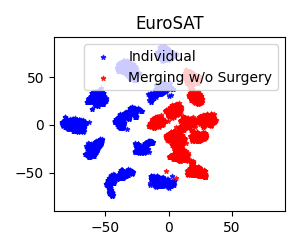}
    \includegraphics[width=.24\textwidth]{images/distribution/lw_adamerging_ViT-B-32_SVHN_finetuned_Merging_wo_Surgery.png}
    \includegraphics[width=.24\textwidth]{images/distribution/lw_adamerging_ViT-B-32_GTSRB_finetuned_Merging_wo_Surgery.png}
    \includegraphics[width=.24\textwidth]{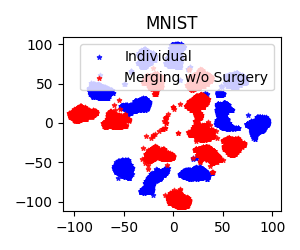}
    \includegraphics[width=.24\textwidth]{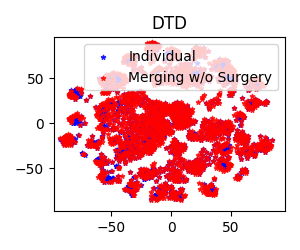}
    % \vspace{-10pt}
     \caption{Visualization of representation distribution of \textbf{AdaMerging (w/o Surgery)} on \textbf{ViT-B/32} architecture.}
\label{fig:distri_vitb32_ladamerging_before} 
\end{figure*}

\begin{figure*}[h]
    \centering 
     \includegraphics[width=.24\textwidth]{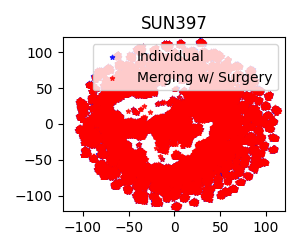}
    \includegraphics[width=.24\textwidth]{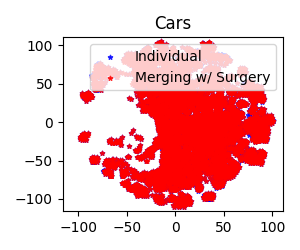}
    \includegraphics[width=.24\textwidth]{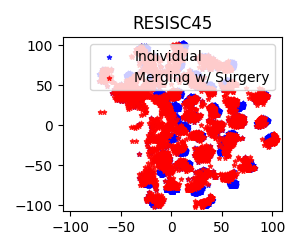}
    \includegraphics[width=.24\textwidth]{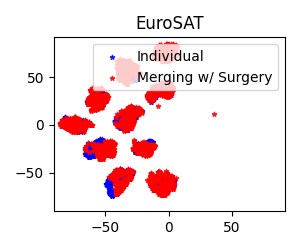}
    \includegraphics[width=.24\textwidth]{images/distribution/lw_adamerging_ViT-B-32_SVHN_finetuned_Merging_w_Surgery.png}
    \includegraphics[width=.24\textwidth]{images/distribution/lw_adamerging_ViT-B-32_GTSRB_finetuned_Merging_w_Surgery.png}
    \includegraphics[width=.24\textwidth]{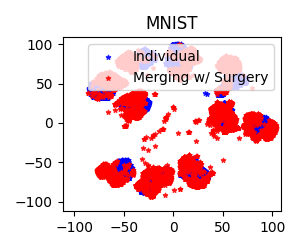}
    \includegraphics[width=.24\textwidth]{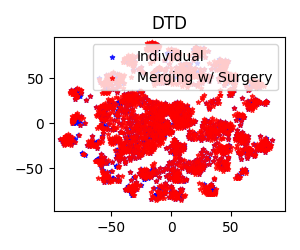}
    % \vspace{-10pt}
     \caption{Visualization of representation distribution of \textbf{AdaMerging (w/ Surgery)} on \textbf{ViT-B/32} architecture.}
\label{fig:distri_vitb32_ladamerging_after} 
\end{figure*}

\begin{figure*}[h]
    \centering 
    \includegraphics[width=.24\textwidth]{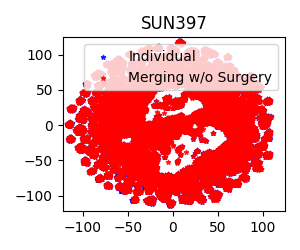}
    \includegraphics[width=.24\textwidth]{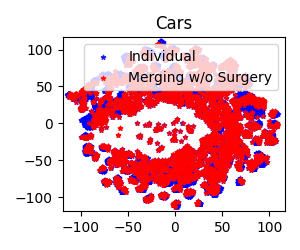}
    \includegraphics[width=.24\textwidth]{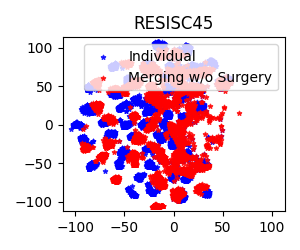}
    \includegraphics[width=.24\textwidth]{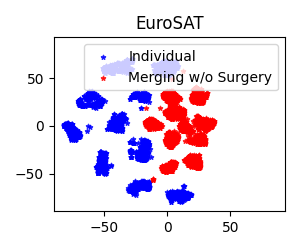}
    \includegraphics[width=.24\textwidth]{images/distribution/lw_adamerging_ViT-B-16_SVHN_finetuned_Merging_wo_Surgery.png}
    \includegraphics[width=.24\textwidth]{images/distribution/lw_adamerging_ViT-B-16_GTSRB_finetuned_Merging_wo_Surgery.png}
    \includegraphics[width=.24\textwidth]{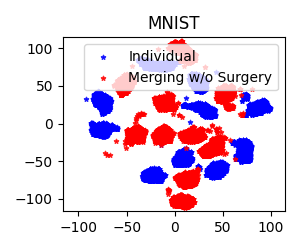}
    \includegraphics[width=.24\textwidth]{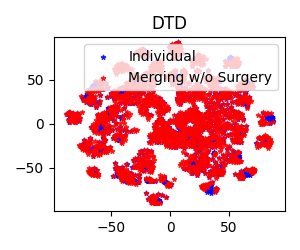}
    % \vspace{-10pt}
     \caption{Visualization of representation distribution of \textbf{AdaMerging (w/o Surgery)} on \textbf{ViT-B/16} architecture.}
\label{fig:distri_vitb16_lwadamerging_before} 
\end{figure*}

\begin{figure*}[h]
    \centering 
     \includegraphics[width=.24\textwidth]{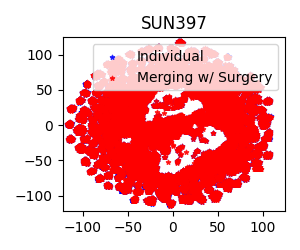}
    \includegraphics[width=.24\textwidth]{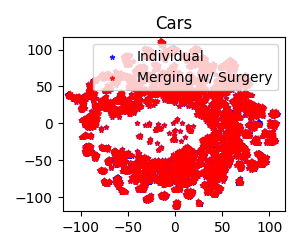}
    \includegraphics[width=.24\textwidth]{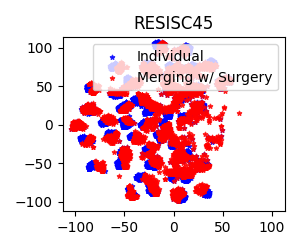}
    \includegraphics[width=.24\textwidth]{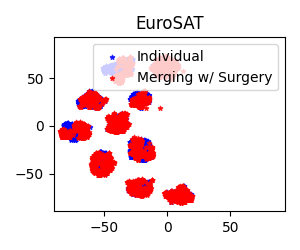}
    \includegraphics[width=.24\textwidth]{images/distribution/lw_adamerging_ViT-B-16_SVHN_finetuned_Merging_w_Surgery.png}
    \includegraphics[width=.24\textwidth]{images/distribution/lw_adamerging_ViT-B-16_GTSRB_finetuned_Merging_w_Surgery.png}
    \includegraphics[width=.24\textwidth]{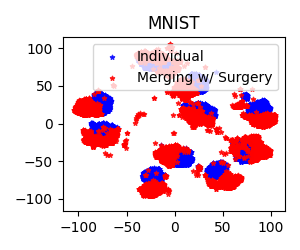}
    \includegraphics[width=.24\textwidth]{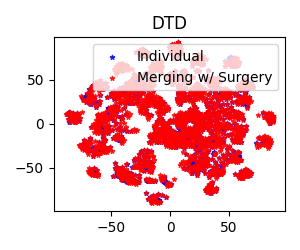}
    % \vspace{-10pt}
    \caption{Visualization of representation distribution of \textbf{AdaMerging (w/ Surgery)} on \textbf{ViT-B/16} architecture.}
\label{fig:distri_vitb16_lwadamerging_after} 
\end{figure*}

\begin{figure*}[h]
    \centering 
    \includegraphics[width=.24\textwidth]{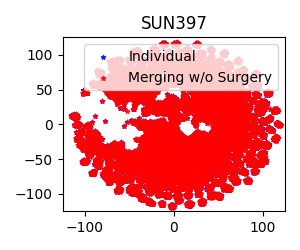}
    \includegraphics[width=.24\textwidth]{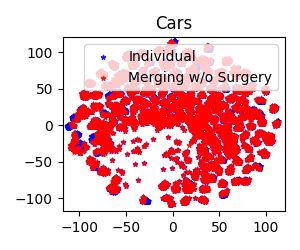}
    \includegraphics[width=.24\textwidth]{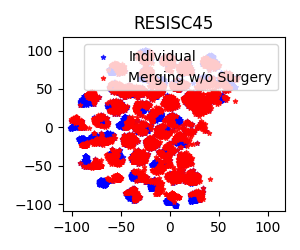}
    \includegraphics[width=.24\textwidth]{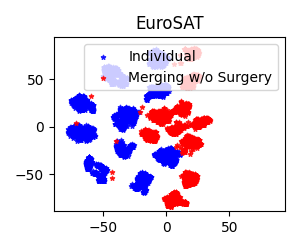}
    \includegraphics[width=.24\textwidth]{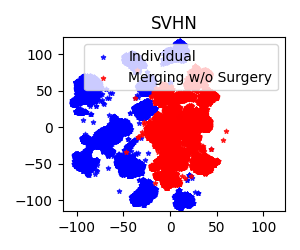}
    \includegraphics[width=.24\textwidth]{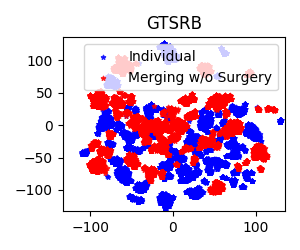}
    \includegraphics[width=.24\textwidth]{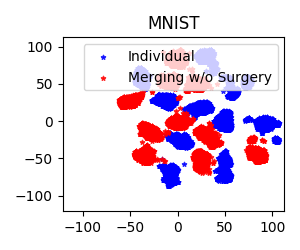}
    \includegraphics[width=.24\textwidth]{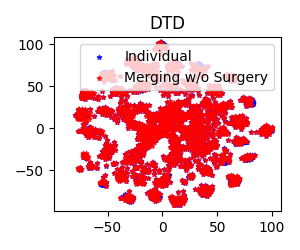}
    % \vspace{-10pt}
    \caption{Visualization of representation distribution of \textbf{AdaMerging (w/o Surgery)} on \textbf{ViT-L/14} architecture.}
\label{fig:distri_vitl14_lwadamerging_before} 
\end{figure*}

\begin{figure*}[h]
    \centering 
     \includegraphics[width=.24\textwidth]{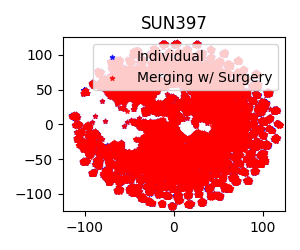}
    \includegraphics[width=.24\textwidth]{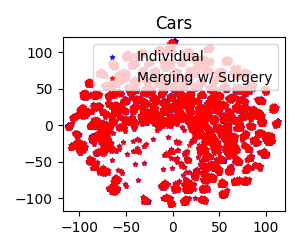}
    \includegraphics[width=.24\textwidth]{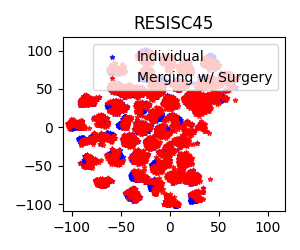}
    \includegraphics[width=.24\textwidth]{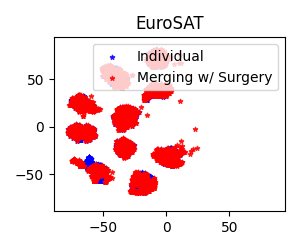}
    \includegraphics[width=.24\textwidth]{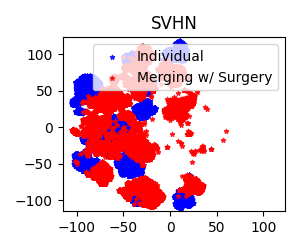}
    \includegraphics[width=.24\textwidth]{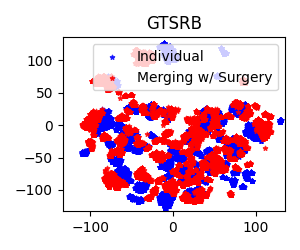}
    \includegraphics[width=.24\textwidth]{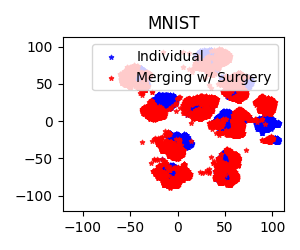}
    \includegraphics[width=.24\textwidth]{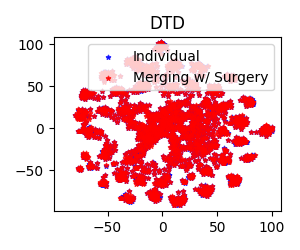}
    % \vspace{-10pt}
     \caption{Visualization of representation distribution of \textbf{AdaMerging (w/ Surgery)} on \textbf{ViT-L/14} architecture.}
\label{fig:distri_vitl14_lwadamerging_after} 
\end{figure*}

\end{document}